%% file: exploring-arxiv.tex
\documentclass[12pt,a4paper]{article}
\usepackage{graphicx}
\usepackage{times}
\usepackage{balance}
\usepackage{url}
\usepackage{caption}
\usepackage{subcaption}

\usepackage{enumitem}
\setenumerate{itemsep=3pt,topsep=0pt,parsep=0pt,partopsep=0pt,leftmargin=1.5pc}

\usepackage{common}
\usepackage{pa}
\usepackage{natbib}

\usepackage{setspace}
\usepackage{titlesec}

\usepackage{endnotes}

\let\footnote=\endnote

\titleformat{\section}
{\singlespacing\normalfont\Large\bfseries}{\thesection}{1em}{}
\titleformat{\subsection}
{\singlespacing\normalfont\large\bfseries}{\thesubsection}{1em}{}
\titleformat{\subsubsection}
{\singlespacing\normalfont\normalsize\bfseries}{\thesubsubsection}{1em}{}

\begin{document}

\title{Exploring the Political Agenda of the European Parliament Using a Dynamic Topic Modeling Approach}

\date{}
\author{Derek Greene\footnote{Insight Centre for Data Analytics \& School of Computer Science, University College Dublin, Ireland (derek.greene@ucd.ie)} \and James P. Cross\footnote{School of Politics \& International Relations, University College Dublin, Ireland (james.cross@ucd.ie).}}

\maketitle

\begin{abstract}
\input{abstract}
\end{abstract}

\clearpage
\doublespacing
\input{intro}

\input{related}
\input{methods}
\input{data}
\input{coherence}

\input{results}

\input{conclusion}

\vskip 1.5em
\noindent {\bf{{Acknowledgement.}}} This research was partly supported by Science Foundation Ireland (SFI) under Grant Number SFI/12/RC/2289.

\input{appendix}


\newpage

\theendnotes

\newpage
\bibliographystyle{chicago}

\bibliography{exploring-arxiv}

\end{document}

%% file: abstract.tex
This study analyzes the political agenda of the European Parliament (EP) plenary, how it has evolved over time, and the manner in which Members of the European Parliament (MEPs) have reacted to external and internal stimuli when making plenary speeches. To unveil the plenary agenda and detect latent themes in legislative speeches over time, MEP speech content is analyzed using a new dynamic topic modeling method based on two layers of Non-negative Matrix Factorization (NMF). This method is applied to a new corpus of all English language legislative speeches in the EP plenary from the period 1999-2014. Our findings suggest that two-layer NMF is a valuable alternative to existing dynamic topic modeling approaches found in the literature, and can unveil niche topics and associated vocabularies not captured by existing methods. Substantively, our findings suggest that the political agenda of the EP evolves significantly over time and reacts to exogenous events such as EU Treaty referenda and the emergence of the Euro-crisis. MEP contributions to the plenary agenda are also found to be impacted upon by voting behaviour and the committee structure of the Parliament. 

%% file: intro.tex
\section{Introduction}
\label{sec:intro}

The plenary sessions of the European Parliament (EP) are one of the most important arenas in which European representatives can air questions, express criticisms and take policy positions to influence European Union (EU) politics. The plenary thus represents the most visible venue where the content and evolution of the policy agenda of the EP can be examined. As a result, understanding how Members of the European Parliament (MEPs) express themselves in plenary, and investigating how the policy agenda of the EP evolves and responds to internal and external stimuli is a fundamentally important undertaking.

In recent years, there has been a concurrent explosion of online records capturing MEP speeches, and the development of data-mining techniques capable of extracting latent patterns in content across sets of these speeches. This allows us for the first time to investigate the plenary agenda of the EP in a holistic and rigorous manner. One approach to tracking the political attention of political figures has been to apply topic-modeling algorithms to large corpora of political texts, such as parliamentary speeches of the U.S. Senate \citep{quinn2010analyze}. These algorithms seek to distill the latent thematic patterns in a corpus of speeches \citep{blei03lda}, and can be used to improve the transparency of the policy agenda by providing a macro-level overview of the political debate in a time- and resource-efficient manner. 

This study takes up the challenge of extracting latent thematic patterns in political speeches by developing a dynamic topic model\footnote{A Python implementation of the proposed dynamic topic model approach is available online: \url{https://github.com/derekgreene/dynamic-nmf}} to investigate how the plenary agenda of the EP has changed over three parliamentary terms (1999--2014). The method applies two layers of Non-negative Matrix Factorization (NMF) topic modeling \citep{lee99nmf} to a corpus of 210,247 speeches from 1,735 MEPs across the 28 EU member states in the EU during that period.

Our proposed topic modeling methodology reveals the breadth of the policy agenda discussed by MEPs in the EP, and the results presented in \refsec{sec:eval} indicate that the agenda has evolved significantly over time. By examining a number of  case studies, ranging from the Euro-crisis to EU treaty changes, we identify the relationship between the evolution of these dynamic topics and the exogenous events driving them. By using external data sources, we can also confirm the semantic and construct validity of these topics. In order to explain some of the patterns we observe in speech making, we conclude the study with an exploration of the determinants of MEP speech-making behavior on the detected topics.\footnote{To provide access to the results of the project to interested parties, we make a browsable version available online: \url{http://erdos.ucd.ie/europarl}}.
Our results relate to the burgeoning literature on political attention, agenda formation, and agenda diversity \citep{baumgartner2009punctuated,downs1972up,jones2005politics,jennings2011effects}

%% file: related.tex
\section{Related Work}
\label{sec:related}

Major efforts to track and explain policy agendas have been developed in recent years. Beginning in the early 1990s, the Policy Agendas Project (PAP) and the Comparative Agendas Project (CAP) have tracked policy agendas across different political systems, including the EU. The major claim in both of these projects is that the variation in the attention that political figures pay to different issues across time can be described by a \textit{punctuated equilibrium} dynamic, whereby issue attention is stable for long periods of time, but these periods are punctuated by short bursts of increased attention \citep{baumgartner1993agendas}. The sudden punctuations in political attention have been explained by factors including the bounded rationality of the political figures involved \citep{jones1994reconceiving}, (re-)framing of policy choices \citep{jones2005politics}, and the influence of exogenous shocks on political priorities \citep{jones2012there,john2012policy}, all of which lead to abrupt spikes in issue attention. Despite some conceptual and measurement challenges \citep{dowding2015comparative}, evidence for the existence of this type of agenda dynamic is found across a multitude of political systems \citep{baumgartner2009punctuated}.

In the EU context, and building upon the techniques developed by the PAP/CAP to capture the aforementioned punctuated-equilibrium dynamic, most academic work has focused on the evolving policy agenda of the European Council \citep{alexandrova2012policy}. Similar to what has been found in other contexts, a punctuated equilibrium dynamic appears to be in play in the European Council, with long periods of agenda stability interrupted with sharp spikes in issue attention. Institutional, contextual and issue-specific factors are found to explain these punctuations. To date, the policy agendas of other EU institutions have been neglected due to the challenges associated with capturing the diverse, diffuse, and multifaceted nature of the policy agendas found in institutions like the Commission, Council of Ministers, and EP.

Despite the fact that policy agenda dynamics in the EP have to date been under-explored, MEP behaviour within the Parliament has been well studied. The most prominent forms of MEP behavior to receive academic attention are plenary speeches and roll-call voting, both of which can be expected to affect the EP policy agenda. Political institutions have been found to shape these forms of MEP behavior. For instance, the formal committee structure of the EP has been shown to provide committee members with strategic advantages due to privileged access to information, and opportunity to shape the EP's policy choices. This has led MEPs to self-select into committees dealing with salient issues with a view to influencing policy outcomes of interest to them \citep{bowler1995organizing}. Within committees, holding roles such as the Chair and Rapporteur have also been shown to affect speech-making and voting behavior \citep{hix2007democratic}. 

Strict institutional rules also govern the allocation of MEP speaking time in the EP plenary \citep{proksch2010position}. The total amount of speaking time for any particular issue is limited and divided between time reserved for actors with formal plenary duties such as rapporteurs, and time proportionally divided between party groups based upon their share of MEPs elected. Speaking time limits lead to competition between MEPs, and party-group leaders allocate scarce speaking time between MEPs for maximum impact \citep{Slapin:2010il}. 

MEP speech content has been shown to reflect latent ideological conflict between MEPs \citep{Slapin:2010il}. Using text-analysis techniques based upon word-frequency distributions, these authors demonstrate the correspondence between the content of legislative speeches and other measures of ideological positions found in the literature based upon roll-call votes and expert surveys. To our knowledge, topic models have yet to be applied to the EP plenary.

Topic models aim to discover the latent semantic structure or topics within a text corpus, which can be derived from co-occurrences of words across documents. These models date back to the early work on latent semantic indexing by \citet{deerwester90lsi}, which proposed the decomposition of term-document matrices for this purpose using Singular Value Decomposition. Considerable research on topic modeling has focused on the use of probabilistic methods, where a topic is viewed as a probability distribution over words, with documents being mixtures of topics, thus permitting a topic model to be considered a generative model for documents \citep{steyvers06prob}. The most widely-applied probabilistic topic modeling approach is Latent Dirichlet Allocation (LDA) proposed by \citet{blei03lda}. Following on from static LDA methods, authors have subsequently developed analogous probabilistic approaches for tracking the evolution of topics over time in a sequentially-organized corpus of documents, such as the dynamic topic model (DTM) of \citet{blei06dynamic}. 

Alternative algorithms, such as Non-negative Matrix Factorization (NMF) \citep{lee99nmf}, have also been effective in discovering the underlying topics in text corpora \citep{wang12group}.  NMF is an unsupervised approach for reducing the dimensionality of non-negative matrices, which seeks to decompose the data into factors that are constrained so as to not contain negative values. By modeling each object as the additive combination of a set of non-negative basis vectors, a readily interpretable clustering of the data can be produced without requiring further post-processing. When working with text data, these clusters can be interpreted as topics, where each document is viewed as the additive combination of several overlapping topics. One of the advantages of NMF methods over existing LDA methods is that there are fewer parameter choices involved in the modelling process. Another advantage that is particularly useful for the application presented in this paper is that NMF is capable of identifying niche topics that tend to be under-reported in traditional LDA approaches (O’Callaghan et al. 2015). In the context of the EP plenary, this is an especially useful attribute of a topic model, as discussions are likely to include a mixture of broader general topics and more specific topics with specialized vocabularies, given the technocratic nature of some EU politics.

Topic-modeling methods have been adopted in the political science literature to analyze political attention. In settings where politicians have limited time-resources to express their views (\eg plenary sessions in parliaments), they must decide which topics to address. Analyzing what they choose to speak about can thus provide insight into the political priorities of the politicians under consideration. Single-membership topic models, which assume each speech relates to one topic, have successfully been applied to plenary speeches made in the U.S. Senate in order to trace political attention of the Senators over time \citep{quinn2010analyze}. This study found that a rich political agenda emerged, where topics evolved over time in response to both internal and external stimuli.

Bayesian hierarchical topic models have also been used to capture the political priorities expressed in Congressional press releases \citep{grimmer10bayesian}, and structural topic models have been used to incorporate text ``metadata" in the form of document-level covariates. Such covariates can include information about a document itself such as when and where it was created, alongside information about the creator of the document \citep{roberts2014structural}.

In conclusion, the current literature provides some interesting insights into the factors that affect MEP speech-making and voting behavior, and the introduction of topic models to the study of political agendas has allowed researchers to consider larger and more complete datasets of political activity across longer time periods than has previously been possible.

%% file: methods.tex
\section{Methods}
\label{sec:methods}

In this section we describe a two-layer strategy for applying topic modeling in a non-negative matrix factorization framework to a timestamped corpus of political speeches. We first describe the application of NMF topic modeling to a single set of speeches from a fixed time period, and then propose a new approach for combining the outputs of topic modeling from successive time periods to detect a set of \emph{dynamic topics} that span part or all of the duration of the corpus. 

\subsection{Topic Modeling Speeches}
\label{sec:methods1}

While work on topic models often involves the use of LDA, NMF can also be applied to textual data to reveal topical structures \citep{wang12group}. The ability of NMF to account for how important a word is to a document in a collection of texts, based on weighted term-frequency values, is particularly useful. Specifically, applying a log-based term frequency-inverse document frequency (TF-IDF) weighting factor to the data prior to topic modeling has shown to be advantageous in producing diverse but semantically coherent topics which are less likely to be represented by the same high-frequency terms. This makes NMF suitable when the task is to identify both broad, high-level groups of documents, and niche topics with specialized vocabularies \citep{ocallaghan15eswa}. In the context of political speech in parliaments, this is a particularly desirable attribute of the model, as it can differentiate between broad procedural topics relating to the day-to-day running of plenary and more focused discussions on specific policy issues. This claim is demonstrated concretely in the analysis below.

\subsubsection{Applying NMF}
\label{sec:nmf}

Given a corpus of $n$ speeches, we first construct a document-term matrix $\m{A} \in \Real^{n\times m}$, where $m$ is the number unique terms present across all speeches (\ie the corpus vocabulary). Applying NMF to $\m{A}$ results in a reduced rank-$k$ approximation in the form of the product of two non-negative factors $\m{A} \approx \m{W}\m{H}$, where the objective is to minimize the reconstruction error between $\m{A}$ and $\m{W}\m{H}$. The rows of the factor $\m{H} \in \Real^{k\times m}$ can be interpreted as $k$ topics, defined by non-negative weights for each of the $m$ terms in the corpus vocabulary. Ordering each row provides a topic descriptor, in the form of a ranking of the terms relative to the corresponding topic. Essentially, the ordered row entries of the matrix $\m{H}$ allow us to identify the most common terms characterizing each topic, thus allowing for substantive interpretation. The columns in the matrix $\m{W} \in \Real^{n\times k}$ provide membership weights for all $n$ speeches with respect to each of the $k$ topics. The columns in matrix $\m{W}$ can be used to associate individual speeches with the topic they are related to, and when we know from meta-data what MEP makes a given speech, we can thus capture MEP contributions to a given topic.

NMF algorithms are often initialized with random factors, which can lead to unstable results where the algorithm converges to a variety of local minima of poor quality. To improve the quality of the resulting topics, we generate initial factors using the Non-negative Double Singular Value Decomposition (NNDSVD) initialization approach \citep{bout08headstart}.

\subsubsection{Parameter Selection}
\label{sec:param}

A key parameter selection decision in topic modeling pertains to the number of topics $k$. Choosing too few topics will produce results that are overly broad, while choosing too many will lead to many small, highly-similar topics. One general strategy proposed in the literature has been to compare the \emph{topic coherence} of topic models generated for different values of $k$ \citep{chang09tea}. A range of such coherence measures exists in the literature, although many of these are specific to LDA. Recently, \citet{ocallaghan15eswa} proposed a general measure, Topic Coherence via Word2Vec (TC-W2V), which evaluates the relatedness of a set of top terms describing a topic. This approach uses the increasingly popular \emph{word2vec} tool \citep{mikolovEfficient} to compute a set of vector representations for all of the terms in a large corpus. We can assess the extent to which the two corresponding terms share a common meaning or context (\eg are related to the same topic) by measuring the similarity between pairs of term vectors. Topics with descriptors consisting of highly-similar terms, as defined by the similarity between their vectors, should be more semantically coherent.

For the purpose of assessing the coherence of topic models, TC-W2V operates as follows. The coherence of a single topic $t_h$ represented by its $t$ top ranked terms is given by the mean pairwise cosine similarity between the $t$ corresponding term vectors in the \emph{word2vec} space:

\begin{equation}
	\textrm{coh}(t_h) = \frac{1}{\binom{t}{2}}\sum_{j=2}^{t} \sum_{i=1}^{j-1} cos(wv_i, wv_j)
	\label{eqn:coh}
	\end{equation} 
		An overall score for the coherence of a topic model $T$ consisting of $k$ topics is given by the mean of the individual topic coherence scores:
	\begin{equation}
	\textrm{coh}(T) =  \frac{1}{k} \sum_{h=1}^{k}\textrm{coh}(t_h) 
	\label{eqn:meancoh}
\end{equation} 

An appropriate value for $k$ can be identified by examining a plot of the mean TC-W2V coherence scores for a fixed range [$k_{min},k_{max}$] and selecting a value corresponding to the maximum coherence.
	
\subsection{Dynamic Topic Modeling}
\label{sec:methods2}

\subsubsection{Layer 1}

When applying clustering to temporal data, authors have often proposed dividing the data into \emph{time windows} of fixed duration \citep{sulo10temporal}. Therefore, following \citet{sulo10temporal}, we divide the full time-stamped corpus of parliamentary speeches into $\tau$ disjoint time windows $\fullset{T}{\tau}$ of equal length. The rationale for the use of disjoint time windows as opposed to processing the full corpus in batch is two-fold: 1) we are interested in identifying the agenda of the parliament at individual time points as well as over all time; 2) short-lived topics, appearing only in a small number of time windows, may be obscured by only analyzing the corpus in its entirety or using overlapping time windows. At each time window $T_i$, we apply NMF with parameter selection based on \reft{eqn:meancoh} to the transcriptions of all speeches delivered during that window, yielding a \emph{window topic model} $M_{i}$ containing $k_{i}$ \emph{window topics}. This process produces a set of successive window topic models $\fullset{M}{\tau}$, which represents the output of the first layer in our proposed methodology.\footnote{It is of course possible to have overlapping time windows that would smooth the transition between time periods in the model. We avoid this specification of the model as it has a smoothing effect which makes it more difficult to identify when topic evolution takes place.}

\subsubsection{Layer 2}

From the window topic models we construct a new condensed representation of the original corpus, by viewing the rows of each factor $\m{H}_{i}$ coming from each window topic model as ``topic documents''. Each topic document contains non-negative weights indicating the descriptive terms for that window topic. We expect that window topics that come from different windows, but share a common theme, will have similar topic documents. We then construct a topic-term matrix $\m{B}$ as follows:

\vskip 0.5em
\begin{enumerate}
	\item Start with an empty matrix $\m{B}$.
	\item For each window topic model $M_{i}$:
	\begin{enumerate}
		\item For each window topic within $M_{i}$, select the $t$ top-ranked terms from the corresponding row vector of the associated NMF factor $\m{H}$, set all weights for all other terms in that vector to 0. Add the vector as a new row in $\m{B}$.
	\end{enumerate}
	\item Once vectors from all topic models have been stacked in this way, remove any columns with only zero values (\ie terms from the original corpus which did not ever appear in the $t$ top ranked terms for any window topics).
\end{enumerate}
\vskip 0.5em

The matrix $\m{B}$ has size $n' \times m'$, where $n' = \sum_{i=1}^{\tau} k_{i}$ is the total number of ``topic documents'' and $m' << m$ is the subset of terms remaining after Step 3. The use of only the top $t$ terms in each topic document allows us to implicitly incorporate feature selection into the process. The result is that we include those terms that were highly descriptive in each time window, while excluding those terms that never featured prominently in any window topic. This reduces the computational cost for the second factorization procedure described below.

Having constructed $\m{B}$, we now apply a second layer of NMF topic modeling to this matrix to identify $k'$ \emph{dynamic topics} that potentially span multiple time windows. The process is the same as that outlined previously, where $\m{B}$ is substituted for the matrix $\m{A}$ when applying NMF as described in \refsec{sec:nmf}. Here the TC-W2V coherence measure is used to detect number of dynamic topics $k'$. The resulting factors $\m{B} \approx \m{U}\m{V}$ can be interpreted as follows: the top ranked terms in each row of $\m{V}$ provide a description of the dynamic topics; the values in the columns of $\m{U}$ indicate to what extent each window topic is related to each dynamic topic.

We track the evolution of these topics over time in the following manner. Firstly, we assign each window topic to the dynamic topic for which it has the maximum weight, based on the values in each row in the factor $\m{U}$. We define the temporal \emph{frequency} of a dynamic topic as the number of distinct time windows in which that dynamic topic appears. The set of all speeches related to this dynamic topic across the entire corpus corresponds to the union of the speeches assigned to the individual time window topics, which are in turn assigned to the dynamic topic.

The resulting outputs of the two-layer topic modeling process are 1) A set of $\tau$ window topic models, each containing $k_{i}$ \emph{window topics}. These are described using their top $t$ terms and the set of all associated speeches; 2) A set of $k'$ \emph{dynamic topics}, each with an associated set of window topics. These are described using their top-$t$ terms and the set of all associated speeches; and 3) A ranking of every MEPs \emph{contributions} relative to all window and dynamic topics in the corpus.

\begin{table}[!t]
	\centering
	\begin{tabular}{|c|p{2.3cm}|p{2.3cm}|p{2.3cm}|p{2.3cm}|}
		\hline\textbf{Rank} & \textbf{2008-Q4} & \textbf{2009-Q1} & \textbf{2009-Q4} & \textbf{2010-Q1} \\\hline
		1             & energy           & climate          & climate       & climate    \\
		2             & climate          & change           & change        & Copenhagen   \\
		3             & emission         & future           & Copenhagen    & change   \\
		4             & package          & emission         & developing    & summit   \\
		5             & change           & integrated       & emission      & emission   \\
		6             & renewable        & water            & conference    & international   \\
		7             & target           & policy           & summit        & Mexico   \\
		8             & industry         & target           & agreement     & conference   \\
		9             & carbon           & industrial       & global         & global  \\
		10            & gas              & global           & energy        & world  \\\hline
	\end{tabular}
	\caption{Example of 4 window topics, described by lists of top 10 terms, which have been grouped together in a single dynamic topic related to climate change.}
	\label{fig:eg-dynamic}
\end{table}

\reftab{fig:eg-dynamic} shows a partial example of a dynamic topic. We observe that, for the four window topics, there is a common theme pertaining to climate change. The evolution of the climate change topic can be seen in the emergence of the terms `Copenhagen', `conference' and `summit' in 2009-Q4 and 2010-Q1, at exactly the time when the Copenhagen climate change summit was underway. Detecting the evolution of topics in this manner is one of the advantages of taking a dynamic approach to capture policy agendas. While the variation across the term lists reflects the evolution of this dynamic topic over the time period (2008-Q4 to 2010-Q1), the considerable number of terms shared between the lists underlines its semantic validity.

%% file: data.tex
\section{Data}
\label{sec:data}

In August 2014 we retrieved all plenary speeches available on Europarl, the official website of the European Parliament, corresponding to parliamentary activities of MEPs during the 5th -- 7th terms of the EP.\footnote{The speech texts used can be found here: \url{http://europarl.europa.eu}.} This resulted in 269,696 unique speeches in 24 languages. While we considered the use of either multi-lingual topic modeling or automated translation of documents, issues with the accuracy and reliability of both strategies lead us to focus on English language speeches in plenary -- either from native speakers or translated -- which make up the majority of the speeches available on Europarl. A corpus of 210,247 English language speeches was identified in total, representing 77.95\% of the original collection. In terms of coverage of speeches from MEPs from the member states, this ranges from 100\% for the United Kingdom, through 87\% for Germany, down to 66.2\% for Romania. However, the most recent state to accede to the EU, Croatia, represents an outlier in the sense that only 2.6\% of speeches were available in English at the time of retrieval due to EP speech translation issues.\footnote{A number of sample-selection issues arise from the variable availability of speeches in English. The first is that MEPs from countries with less speeches in English will be systematically under-represented in the corpus, and our substantive results should be interpreted with this in mind. There is unfortunately little that can be done about this until such time as English translations are made available.}

Following considerable previous work on time-stamped document collections (\eg \citet{blei06dynamic}), we subsequently divided the data into a set of sequential non-overlapping slices or \emph{time windows} -- specifically, 60 quarterly  windows from 1999-Q3 to 2014-Q2. We select a quarter as the time window duration to allow for the identification of granular topics, while also ensuring there exists a sufficient number of speeches in each time window to perform topic modeling. Initial experiments performed on shorter durations with small numbers of speeches per window often yielded results with a smaller number of coherent topics. In addition, a quarterly time window is appropriate in order to avoid empty time windows occurring due to the summer recess of the EP. 

For each time window $T_{i}$ we construct a document-term matrix $\m{A}_{i}$ as follows:
\begin{enumerate}[leftmargin=3em,rightmargin=\leftmargin]
	\item Select all speech transcriptions from window $T_{i}$, and remove all header and footer lines.
	\item Find all unigram tokens in each speech, through standard case conversion, tokenization, and lemmitization.
	\item Remove short tokens with $< 3$ characters, and tokens corresponding to generic stop words (\eg ``are'', ``the''), parliamentary-specific stop words (\eg ``adjourn'', ``comment''), and names of politicians.
	\item Remove tokens occurring in $< 5$ speeches.
	\item Construct matrix $\m{A}_{t}$, based on the remaining tokens. Apply TF-IDF term weighting and document length normalization.
\end{enumerate}

The resulting time window data sets range in size from 679 speeches in 2004-Q3 to 9,151 speeches in 2011-Q4, with an average of 4,811 terms per data set. 

%% file: coherence.tex
\section{Assessing the Coherence of LDA and NMF\\ Topic Models: A Baseline Comparison}
\label{sec:coh}

As noted previously, probabilistic methods such as LDA have been widely applied for topic modeling, although recent work has shown that factorization-based algorithms such as NMF are effective in identifying niche topics with more specific vocabularies \citep{ocallaghan15eswa}. In the context of EU politics where technocratic issues are often discussed, being able to detect these niche topics should be an advantage. To illustrate this idea, Table \ref{tab:LDAcrisis} shows the top-5 terms associated with a topic relating to the Eurocrisis for a selection of consecutive time window as produced by competing NMF and LDA approaches. Terms in bold are unique to a topic produced by a given approach, while terms in italics are found in both sets of terms. As can be seen, the vocabulary produced by NMF to describe the Eurocrisis are much more rich and varied compared to those produced by LDA. If we were interested in the content and dynamics of the debate surrounding the Eurocrisis, then NMF appears to produce a more informative and time-variant picture of debate evolution.

\begin{table}[!ht]
	\centering
	\begin{tabular}{|l|lllll|}\hline
	NMF & Term 1 & Term 2 & Term 3 & Term 4 & Term 5\\\hline
	2010-Q1 & \textit{crisis} & \textit{economic} & \textit{financial} & \textbf{strategy} & \textbf{current} \\
	2010-Q2 & \textit{financial} & \textbf{supervision} & \textit{crisis} & \textit{economic} & \textbf{package} \\
	2010-Q3 & \textit{financial} & \textbf{supervision} & \textit{crisis} & \textit{economic} & \textbf{package} \\
	2010-Q4 & \textit{economic} & \textit{crisis} & \textit{financial} & \textit{euro} & \textbf{stability} \\
	2011-Q1 & \textit{financial} & \textbf{tax} & \textbf{pension} & \textbf{system} & \textit{economic} \\
	2011-Q2 & \textbf{surveillance} & \textit{euro} & \textbf{budgetary} & \textit{economic} & \textbf{macroecon} \\
	2011-Q3 & \textit{economic} & \textit{crisis} & \textit{euro} & \textbf{growth} & \textbf{policy} \\
	2011-Q4 & \textit{economic} & \textit{crisis} & \textit{financial} & \textbf{policy} & \textbf{states}\\\hline
	LDA & Term 1 & Term 2 & Term 3 & Term 4 & Term 5\\\hline
	2010-Q1 & \textbf{european} & \textit{crisis} & \textit{financial} & \textit{economic} & \textit{euro} \\
	2010-Q2 & \textbf{european} & \textit{crisis} & \textit{financial} & \textit{economic} & \textit{euro} \\
	2010-Q3 & \textbf{european} & \textit{financial} & \textit{crisis} & \textit{economic} & \textit{euro} \\
	2010-Q4 & \textbf{european} & \textit{crisis} & \textit{financial} & \textit{economic} & \textit{euro} \\
	2011-Q1 & \textbf{european} & \textit{crisis} & \textit{economic} & \textit{financial} & \textit{euro} \\
	2011-Q2 & \textbf{european} & \textit{crisis} & \textit{economic} & \textit{financial} & \textit{euro} \\
	2011-Q3 & \textbf{european} & \textit{crisis} & \textit{economic} & \textit{financial} & \textit{euro} \\
	2011-Q4 & \textbf{european} & \textit{crisis} & \textit{economic} & \textit{euro} & \textit{financial}\\\hline
	\end{tabular}
	\caption{NMF and LDA - Euro Crisis Topic Top-5 terms. \textbf{Bold} terms unique to one set of results, \textit{italic} terms shared.}
	\label{tab:LDAcrisis}
\end{table}

In order to more systematically compare NMF and LDA topic models, we apply topic coherence methods \citep{stevens12coherence} to assess model performance. Topic coherence refers to the level of semantic similarity between the top terms used to represent a topic (\ie the topic descriptors). We apply these measures to each of our $60$ time window data sets described in \refsec{sec:data} for different numbers of topics $k \in [10,50]$. We apply NMF as described in \refsec{sec:nmf} and use LDA as implemented in the MALLET toolkit \citep{mccallum02mallet}, with hyper-parameter values $\alpha=0.01$ and $\beta=50/k$ as recommended by \citet{steyvers06prob}. 
In our experiments, we calculate topic coherence using two different measures to evaluate the top 10 terms for each topic in all 300 models produced by each algorithm (\ie 60 datasets for 5 different values of $k$). Following \citet{roder15palmetto}, we report the median value to provide a more robust score summarising each model.

\begin{figure}[!t]	
	\centering
	\begin{subfigure}[t]{0.485\linewidth}
		\centering
		\includegraphics[width=\linewidth]{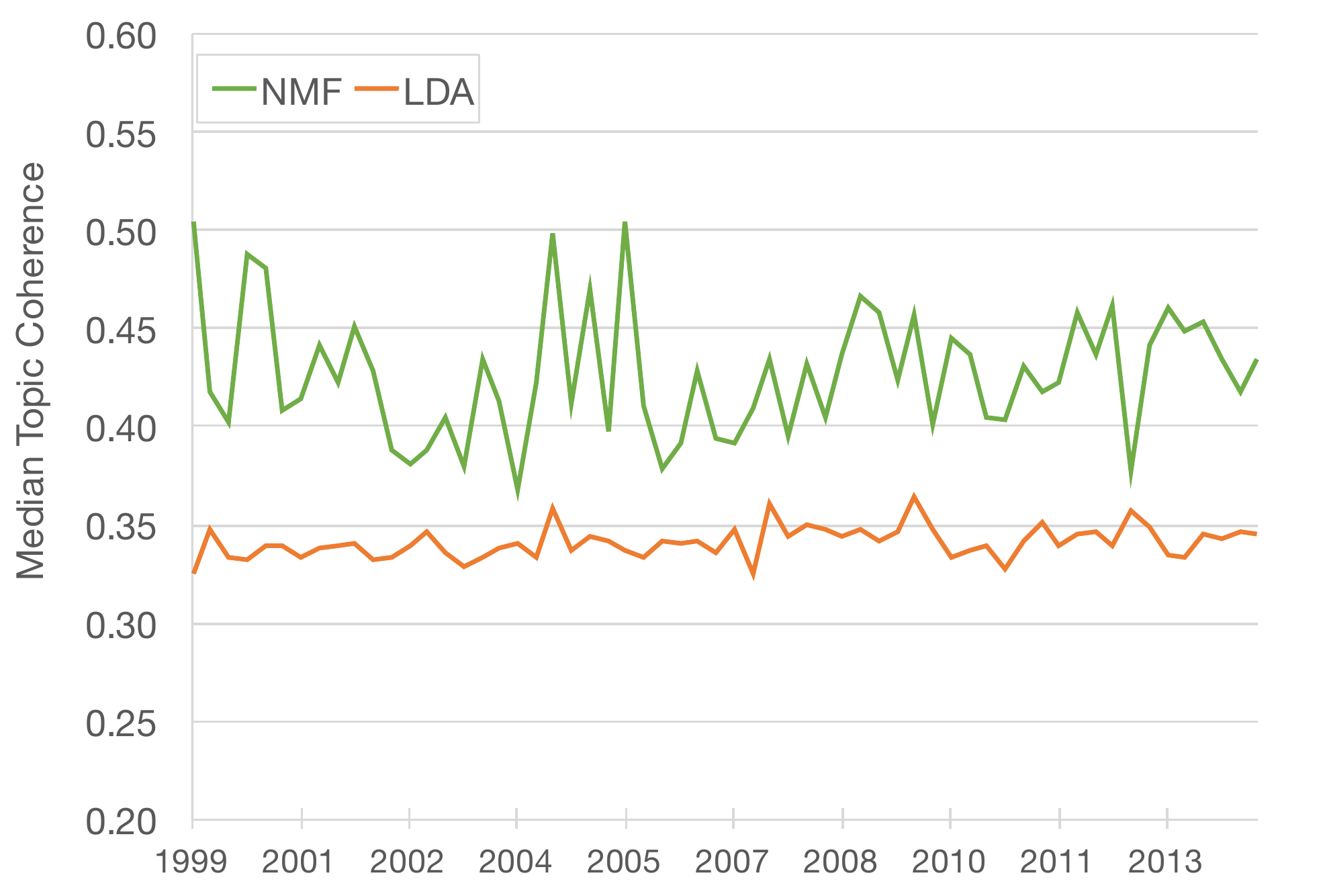}
		\caption{$C_{v}$ at $k=10$ topics}		
		\label{fig:coh1}
	\end{subfigure}
	\hskip 0.5em
	\begin{subfigure}[t]{0.485\linewidth}
		\centering
		\includegraphics[width=\linewidth]{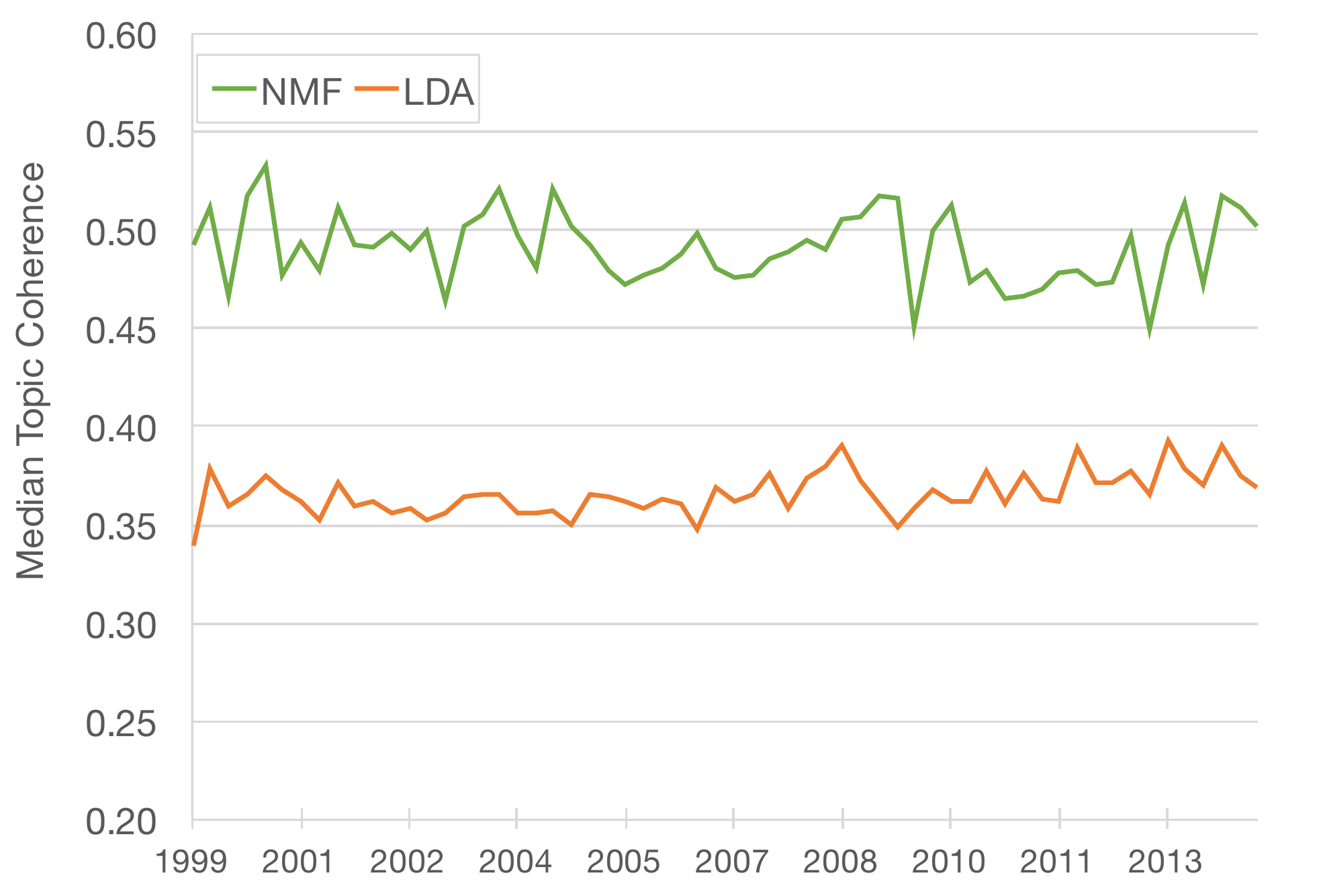}
		\caption{$C_{v}$ at $k=50$ topics}		
		\label{fig:coh2}
	\end{subfigure}
	\vskip 1em
	\begin{subfigure}[t]{0.485\linewidth}
		\centering
		\includegraphics[width=\linewidth]{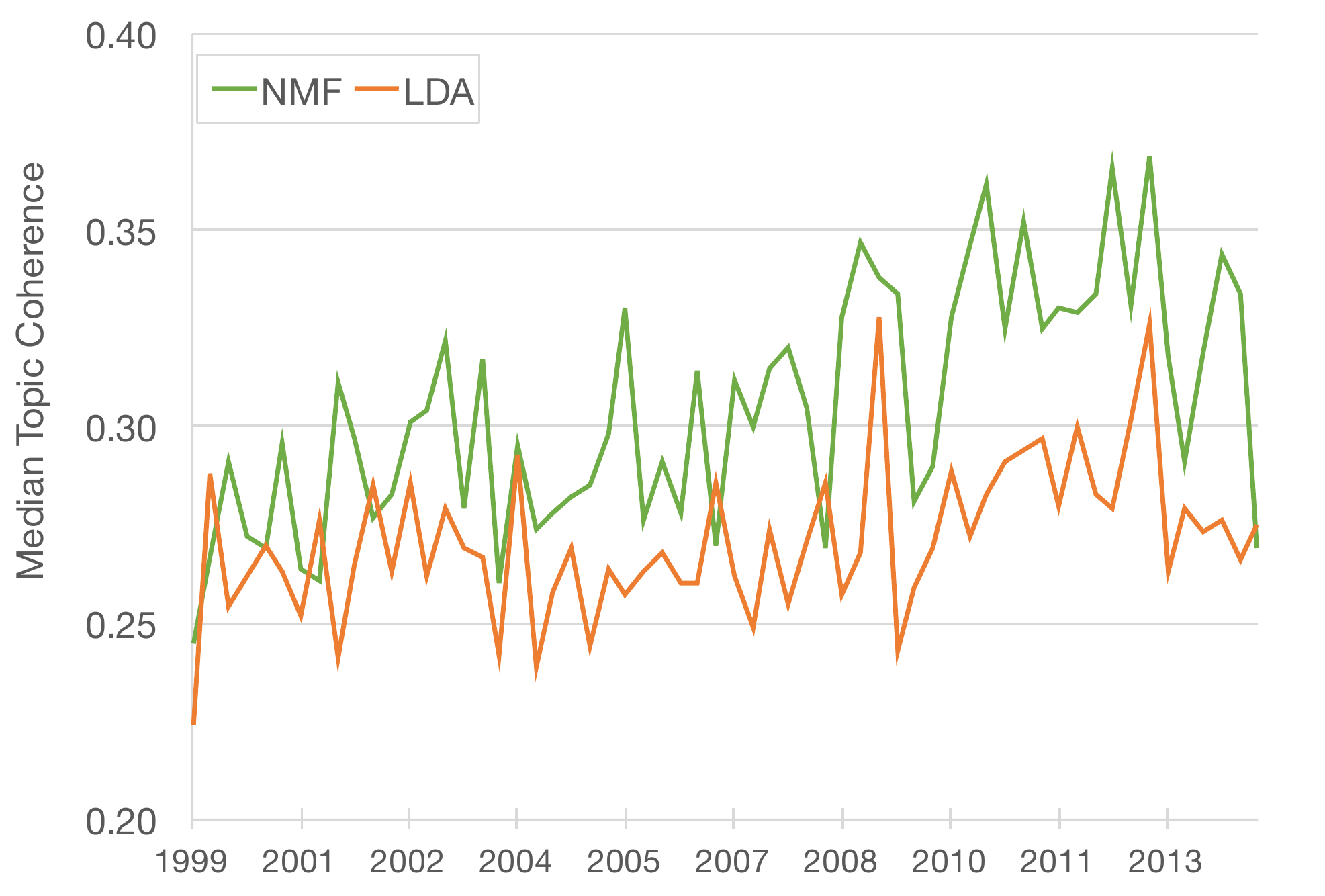}
		\caption{TC-W2V at $k=10$ topics}		
		\label{fig:coh3}
	\end{subfigure}
	\hskip  0.5em
	\begin{subfigure}[t]{0.485\linewidth}
		\centering
		\includegraphics[width=\linewidth]{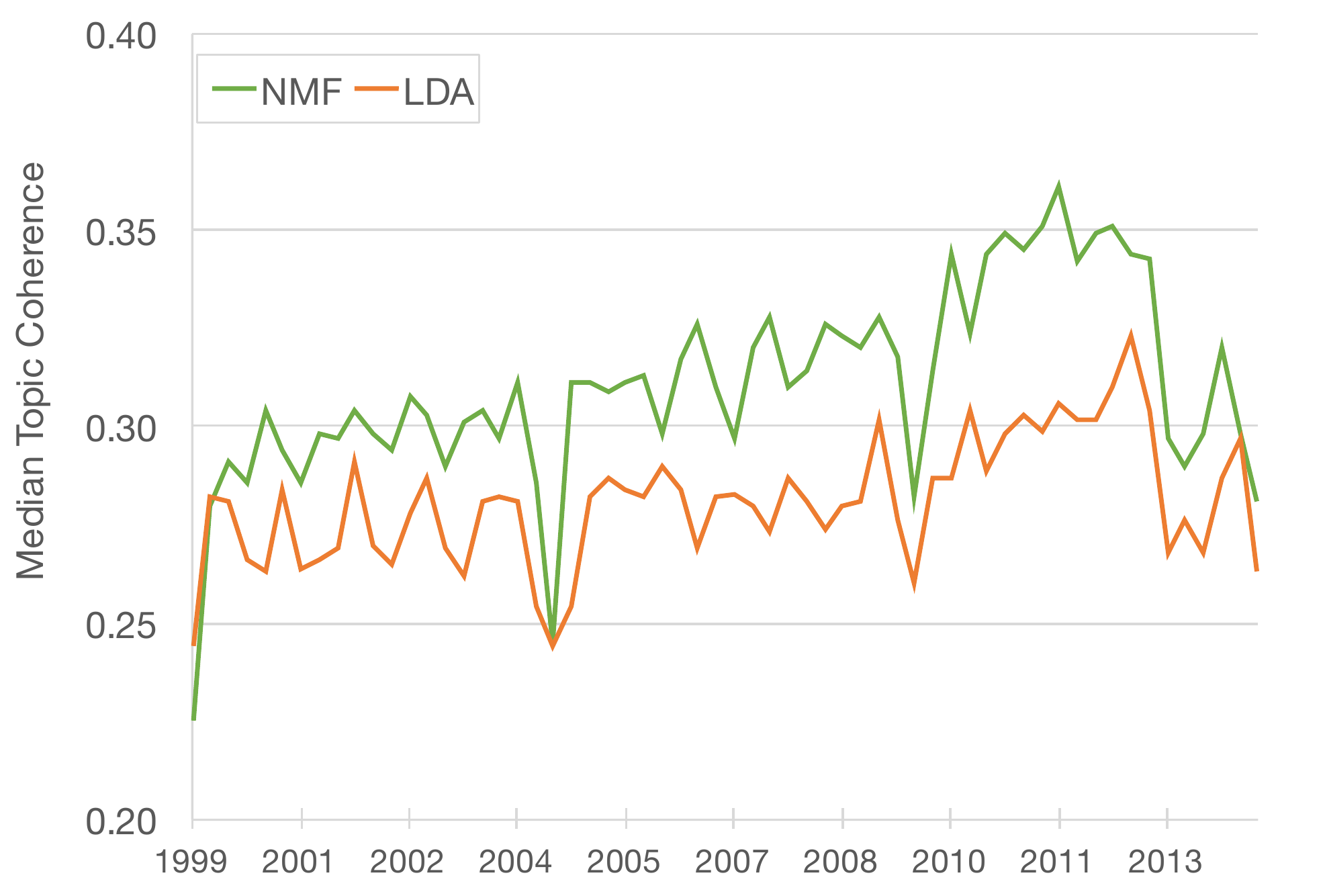}
		\caption{TC-W2V at $k=50$ topics}		
		\label{fig:coh4}
	\end{subfigure}
	\caption{Median topic-coherence scores for all 60 time window datasets, for topic models produced by NMF and LDA, using the $C_{v}$ and TC-W2V measures.}
	\label{fig:coh}
\end{figure}

Firstly, we compare NMF and LDA using the $C_{v}$ topic coherence measure which was identified by \cite{roder15palmetto} as being particularly appropriate for evaluating topic quality, based on a large empirical comparison of different coherence measures. The $C_{v}$ measure retrieves co-occurrence counts from a background corpus for a set of topic terms using a sliding window approach. These counts are then used to calculate the normalized point-wise mutual information (NPMI) for all pairs of terms within a topic descriptor. The intuition behind such an approach is that more coherent topics will have topic descriptors that co-occur more often together across the corpus. As our background corpus for each time-window dataset, we use the full set of 210,247 English EP speeches. Results for $k=10$ and $k=50$ topics are shown in Figures \ref{fig:coh1} and \ref{fig:coh2} respectively. We observed that NMF achieves higher topic-coherence scores across all of the time window data sets for all values of $k$. Inspecting the results suggests that this is largely due to the ability of NMF to uncover more niche and specific topics on the data, compared to the more broad and ultimately less semantically coherent topics extracted by LDA (\ie the term descriptors are not distinctive). 

To further investigate the differing topic coherence resulting from LDA and NMF models, we repeated the process using the TC-W2V coherence measure described previously in \refsec{sec:param}. As described in \refsec{sec:param}, this measure is based on a Word2Vec model, which provides a computationally efficient and effective method for quantifying the semantic relatedness of terms in a large corpus \citep{mikolovEfficient}. Again as our reference corpus we use the full corpus of English speeches. Representative results are shown in Figures \ref{fig:coh3} and \ref{fig:coh4}. As with the $C_v$ measure, we observe that NMF consistently achieves high coherence scores, with a higher score than LDA in the case of 94.7\% of all 300 experiments. We also observe that the TC-W2V measure is generally more sensitive to changes in the top terms used to represent topics, highlighting time windows where topics have a greater or less level of coherence for both algorithms. These results provide our rationale for the use of NMF in the remainder of this paper.

%% file: results.tex
\section{Experimental Results}
\label{sec:eval}

\subsection{Experimental Setup}

\begin{figure}[!t]
	\centering
	\begin{subfigure}{1\textwidth}
		\centering
		\includegraphics[width=0.6\linewidth]{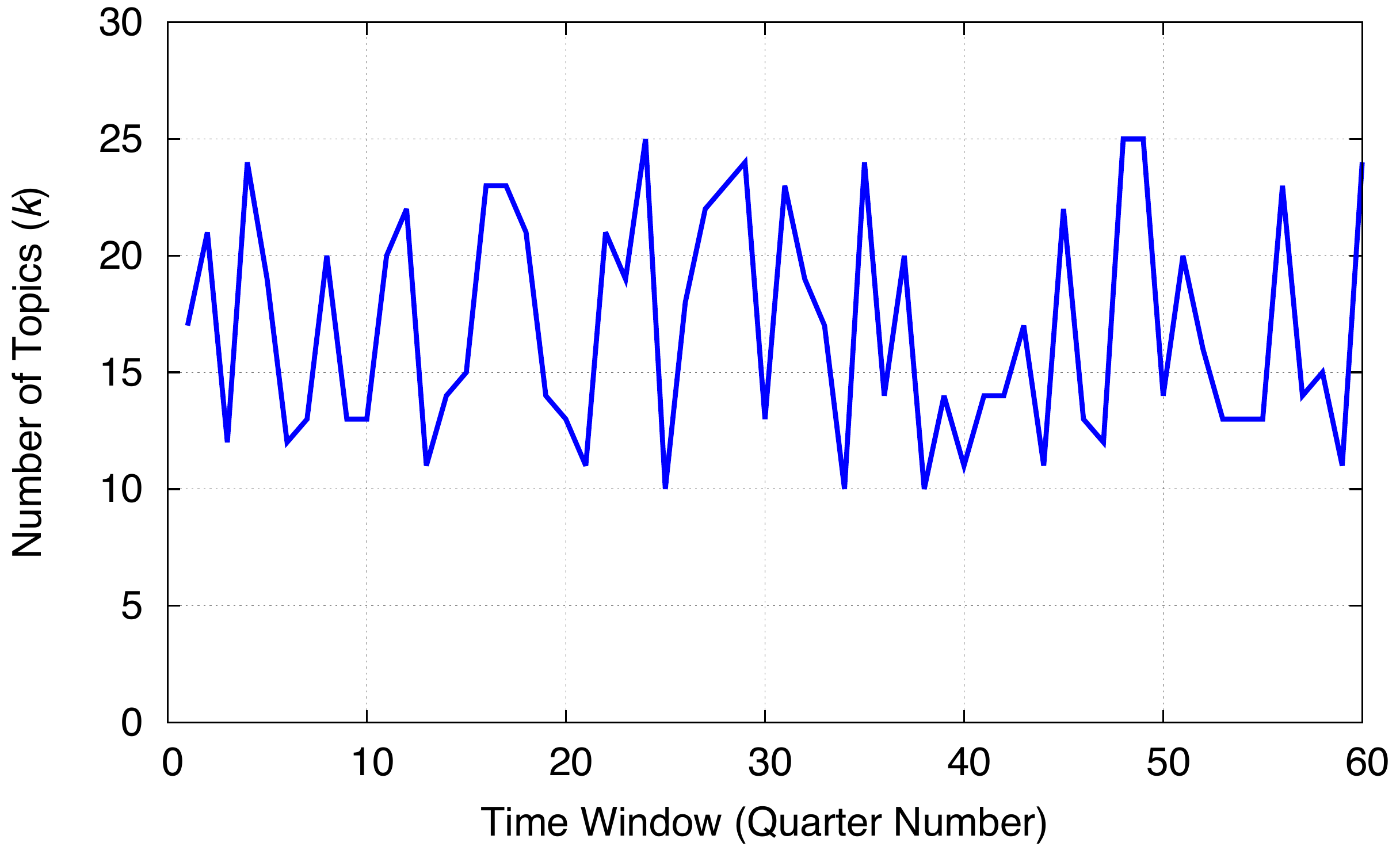}
\caption{Number of window topics identified from 1999-Q3 (\#1) to 2014-Q2 (\#60).}
\label{fig:numtopics}
	\end{subfigure}

	\begin{subfigure}{1\textwidth}
		\centering
		\includegraphics[width=0.6\linewidth]{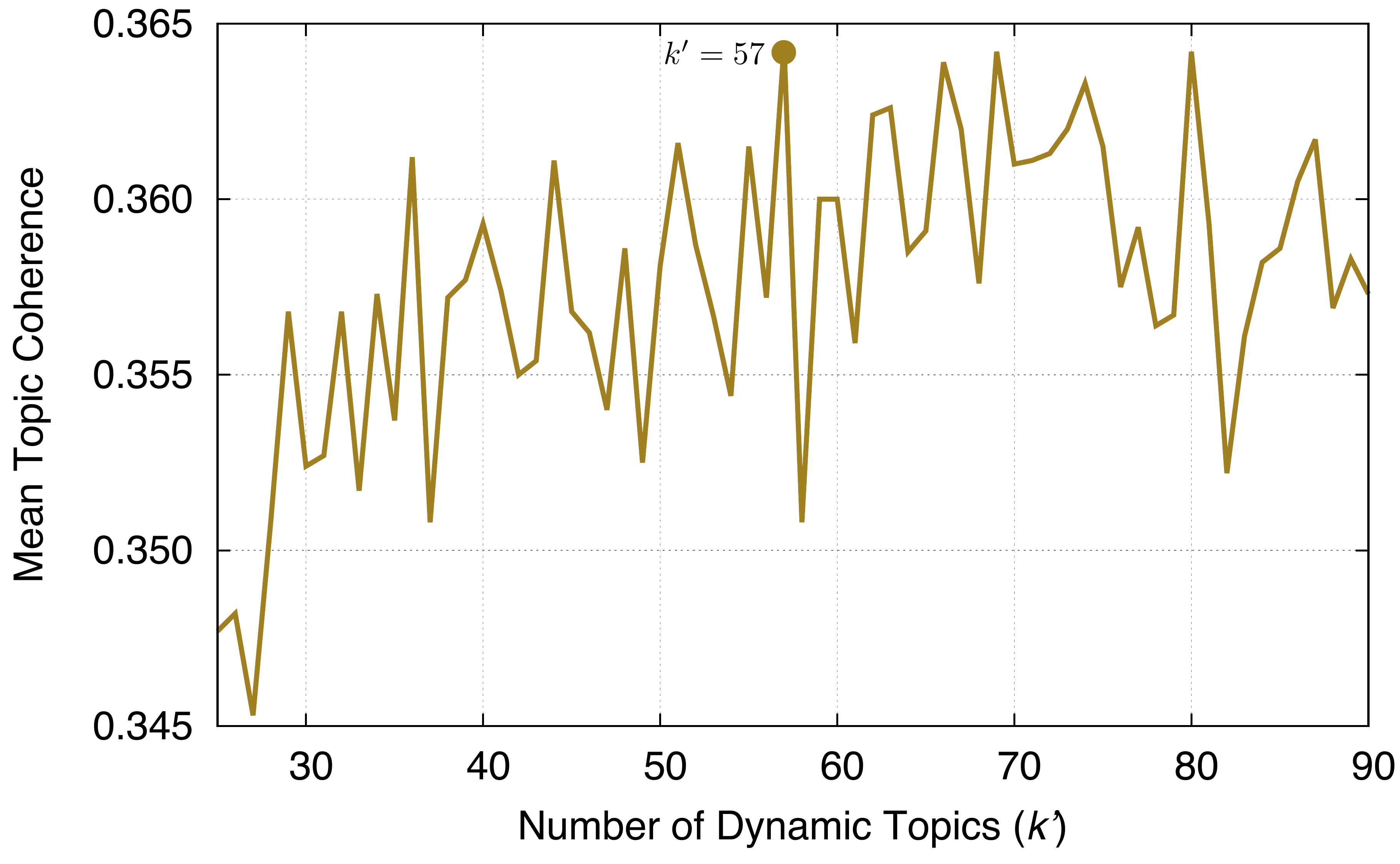}
		\caption{Plot of mean TC-W2V topic-coherence scores for different values for the number dynamic topics of $k'$, across a candidate range $[25,90]$.}
	    \label{fig:numdynamic}
	\end{subfigure}
	\caption{Identifying optimal number of topics using TC-W2V topic coherence.}
	\label{fig:test}
\end{figure}

After pre-processing the data, the first task was to identify $k$, the number of topics in each window. To do this, we applied NMF with parameter selection as described in \refsec{sec:methods1}. Given the relatively specialized vocabulary used in EP debates, when building the \emph{word2vec} space for parameter selection, we used the complete set of English language speeches as our background corpus. We used the same \emph{word2vec} settings and number of top terms per topic ($t=10$) as described in \citet{ocallaghan15eswa}. At each time window, we generated window topic models containing $k \in [10,25]$ topics, and then selected the value $k$ that produced the highest mean TC-W2V coherence score (Eqn. \ref{eqn:meancoh}). The illustration of the number of topics per window in \reffig{fig:numtopics} shows that there is considerable variation in the number of topics detected for each window, which does not correlate with the number of speeches per quarter (Pearson correlation $0.006$). This suggests our results are not driven by the volume of speeches, but rather variation in topics being discussed across different windows.

The process above yielded 1,017 window topics across the 60 time windows. We subsequently applied dynamic topic modeling as described in \refsec{sec:methods2}. For the number of terms $t$ representing each window topic, we experimented with values from 10 to the entire number of terms present in a time window. However, values $t > 20$ did not result in significantly different dynamic topics. Therefore, to minimize the dimensionality of the data, we selected $t=20$. This yielded a matrix of 1,017 window topics represented by 2,710 distinct terms.

Our next task was to identify a value for the parameter $k$ for the dynamic part of the model, i.e.- the number of dynamic topics in the corpus. To do this, we calculated TC-W2V coherence scores for a set of topic models with a range $k' \in [25,90]$ and then compared these coherence scores to identify the appropriate parameter value. The resulting plot (see \reffig{fig:numdynamic}) indicated a maximal value at $k'=57$, although a number of close peaks exist in the range [62,80]. When we manually inspected the results of the most coherent topic models for these values of $k$', they were highly similar in terms of the topics detected, with minor variations corresponding to merges or splits of strongly-related topics.

\input{results_case}
\input{results_explain}

%% file: results_case.tex
\subsection{Case Studies}
\label{sec:case}

\begin{figure}[ht!]
	\centering
	\begin{subfigure}{0.9\textwidth}
		\centering
	\includegraphics[width=0.85\linewidth]{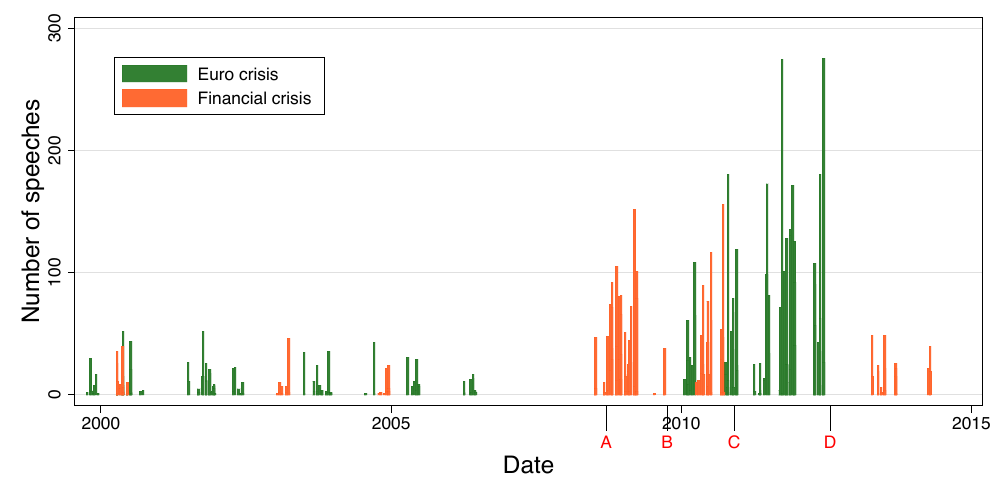}
		\caption{``Financial \& Euro crises'' dynamic topics}
		\label{fig:dynamic1}
	\end{subfigure}
	\begin{subfigure}{0.9\textwidth}
		\centering
		\includegraphics[width=0.85\linewidth]{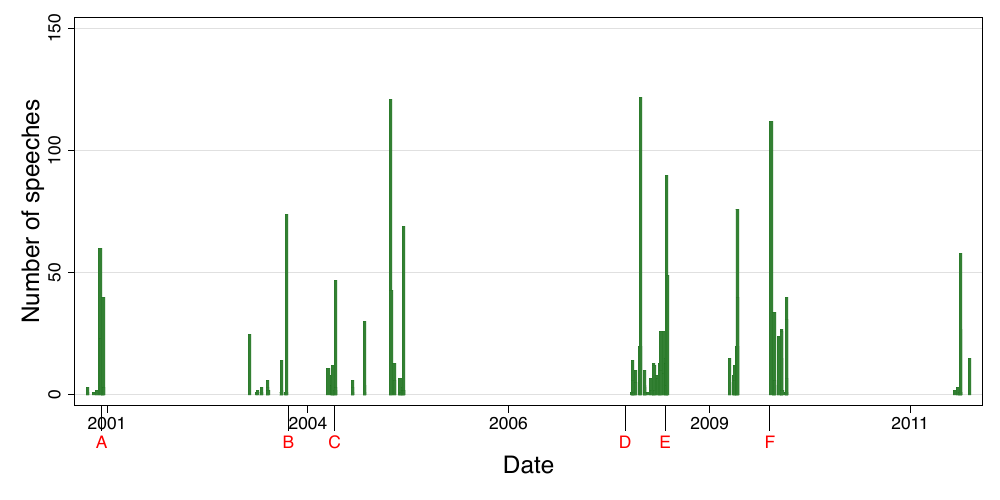}
		\caption{```Treaty changes \& referenda'' dynamic topic}
		\label{fig:dynamic2}
	\end{subfigure}
	\begin{subfigure}{0.9\textwidth}
		\centering
		\includegraphics[width=0.85\linewidth]{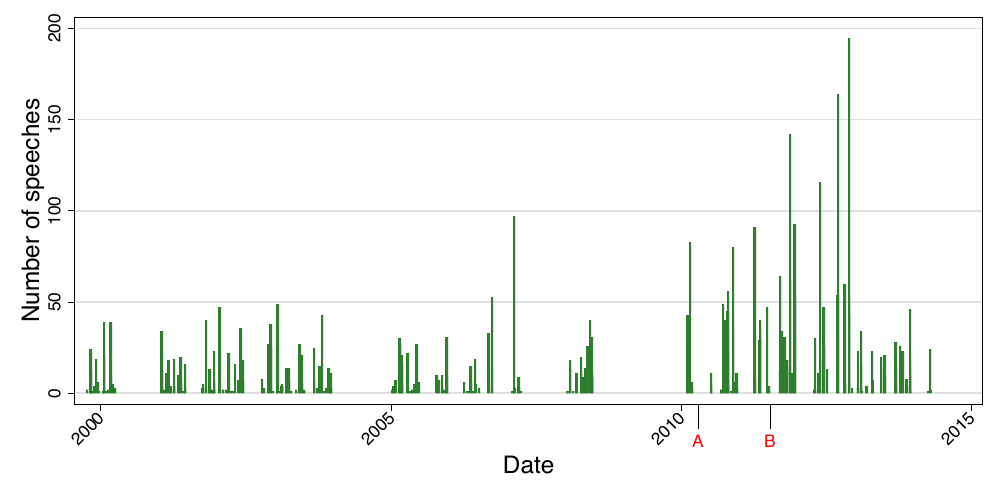}
		\caption{``Fisheries'' dynamic topic}
		\label{fig:dynamic3}
	\end{subfigure}
	\caption{Time plots for three sample dynamic topics across all time windows, from 1999-Q3 (time window \#1) to 2014-Q2 (time window \#60).}
	\label{fig:dynamic}
\end{figure}

In order to investigate the construct validity of our topics, we focus on three case studies to demonstrate how our topic modeling strategy captures variation in the EP policy agenda over time.\footnote{We present extensive topic external and internal validation exercises in the appendix.}

\subsubsection{Financial/Euro-crisis}

Our first case study, illustrated in \reffig{fig:dynamic1}, relates to two topics covering the financial and Euro-crisis respectively. This is an interesting case study, as the initial financial crisis peaked in 2008, and the Euro-crisis that followed has gone through a number of phases starting in 2009. These events can be thought of as exogenous shocks to the policy agenda, and their exogenous nature provides a way to externally validate the dynamic topic modeling approach in use here. \reffig{fig:dynamic1} demonstrates a number of distinct peaks in MEP attention to both the financial crisis topic (in orange) and the Euro-crisis topic (in green). Attention to the financial crisis starts to rise in 2008-Q3 and initially peaks in 2008-Q4 (point A in \reffig{fig:dynamic1}), corresponding to the collapse of the Lehman Brothers bank (15/9/2008). The other peaks in attention in \reffig{fig:dynamic1} correspond to important events in the Euro-crisis. Point B corresponds to the revelations about under-reporting of Greek debt in October 2010, Point C to the Irish bailout (November 2010), and Point D to Mario Draghi's statement that the ECB was ``ready to do whatever it takes to preserve the euro" (July 2012). Draghi's statement temporarily at least reassured markets, thus explaining why fewer speeches relating to this topic are observed after Point D. In effect, it appears that both the financial and Euro crises had the effect of punctuating a rather low-level equilibrium of attention to issues relating to the common currency and financial regulation that existed before 2008 and 2010 respectively.

\subsubsection{Treaty Reform}

Our second case study (\reffig{fig:dynamic2}) relates to EU Treaty reforms. This topic is of interest, because one would expect that MEP attention to the topic varies over time, as Treaty revisions are not common. For example, the Nice Treaty was agreed upon in 2001 and put to a referendum in Ireland in June 2001. The `No' vote that resulted from this referendum accounts for Point A in \reffig{fig:dynamic2}. Similarly, Point B in \reffig{fig:dynamic2} corresponds to the October 2003 Intergovernmental Conference negotiating the Constitutional Treaty. Point C indicates the date the Enlargement Treaty was signed in May 2004. MEP attention relating to the Lisbon Treaty peaks when it was signed (Point D), and when the Irish rejected the Treaty in June 2008 (Point E). Point F corresponds to the second Irish referendum approving Lisbon in October 2009. If we view this variation in attention to Treaties in light of punctuated equilibrium theory, it would appear that equilibrium levels of attention to Treaty changes in the EP is low, but this equilibrium is disturbed with spikes in attention when major exogenous events relating to Treaty change occur. The EP appears to be reactive rather than proactive in this regard (attention spikes after an event), which is not surprising given its limited formal role in Treaty negotiations.

\subsubsection{Fisheries Policy}

Our final case study relates to fisheries policy. Fisheries is an interesting policy-agenda item for the dynamic topic modeling approach to detect, because it relates to the day-to-day functioning of the EU as a fisheries industry regulator, rather than more headline-making policies and events already discussed. As a result one would expect a more constant level of attention to this agenda item with fewer punctuations. \reffig{fig:dynamic3} demonstrates the prevalence of the fisheries topic over time. As can be seen, MEPs pay a reasonably stable level of attention to fisheries between 2000 and 2010. This trend is interrupted in 2010, when MEP attention to fisheries increases. This corresponds to the Commission launching a public consultation on reforming EU fisheries policy in 2009, the results of which were presented to the EP in April 2010. Point A corresponds to the launch of this working document, while Point B corresponds with Commissioner Maria Damanaki introducing a set of legislative proposals designed to reform the common fisheries policy in a speech to the EP in July 2011. This is highly consistent with the patterns in agenda change described in punctuated equilibrium theory.

\vskip 0.5em
In general, the fact that the variation over time that we observe in MEP attention to these case-study topics appears to be driven by exogenous events provides a form of construct validity for our topic modeling approach, and support for the idea that political agendas are relatively stable, but experience punctuations due to exogenous events \citep{john2012policy,jones2012there}. 

%% file: results_explain.tex
\subsection{Explaining MEP Speech Counts}
\label{sec:explain}

We now focus our attention on the 7th EP term that sat between 2009 and 2014, as a set of interesting covariates are available at the MEP level that can help us explain MEP contributions to a given agenda item. We aim to explore the determinants of MEP topic contributions with reference to existing theories in the literature that show MEP ideology, party membership and institutional structure affect other forms of MEP behaviour (including propensity to speak in plenary, rebelling against party principals, and report writing). Our first dependent variable is constructed directly from the dynamic topic model level-2 speech-topic weight matrix, in which each MEP speech can relate to multiple topics. The variable is simply the sum of all weights per MEP for each topic across the entire EP7 term, and thus captures the relative contribution of each MEP to each topic if one assumes that speeches can relate to multiple topics. The skewed and continuous nature of the variable being examined implies that a generalized linear model from the Gaussian family with a log-link function is appropriate for our analysis.

In constructing our second dependent variable, we assume that each speech belongs to one topic alone by allocating each speech to a topic based on the maximum topic weight observed for that speech.\footnote{We also experimented with other alternatives for allocating speeches to topics including allocating a speech to a topic if it is in any way related to a topic (i.e. any non-zero weight observation in the speech-topic matrix counts), and allocating any speech with an above-average weight in the topic window to a given topic. Both of these measures are problematic as they make strong assumptions about very small NMF weights indicating a speech is fully relevant to a topic.} Substantively it is reasonable to assume single-topic memberships for each speech, given that MEP speaking time is limited, thus focusing MEP attention on particular agenda items rather than allowing speeches addressing multiple issues. Making such an assumption has the advantage of providing a more readily interpretable analysis, as the resulting variable captures each MEP's contribution to each topic in terms of a speech count rather than a sum of NMF weights. We employ a negative binomial regression model suitable for analyzing count data with over-dispersion on the resulting variable \citep{cameron2013regression}. In both Model 1 and Model 2 we cluster standard errors by MEP.\footnote{Alternative model specifications including a zero-inflated model were also experimented with, with similar results. We present the negative binomial model here as it is the simpler model and the substantive results are broadly similar to these other models.}

In order to explain the variation observed in our dependent variables, we include independent variables relating to MEP's ideology, voting behavior, and the institutional structures in which they find themselves embedded within, as these variables have been found to be relevant to speech-making behaviour in the EP \cite{proksch2014politics}. We account for the left-right ideological position of an MEP's national party (as a proxy for MEP ideology) using data from \cite{scully2012national}. Following \cite{proksch2014politics}, we also include a measure of how often MEPs vote against their European party group in favor of their national party and vice versa. The idea behind including these variables is that MEPs rebelling against one party affiliation in favor of another will either try to explain such behavior in their speeches thus increasing the observed speech count, or hide their behavior by making no speeches, thus decreasing the observed count. These data were taken from an updated version of the \cite{hix2006dimensions} dataset provided by those authors. In order to capture an MEP's committee positions we include dummies for committee membership, chairs, and Rapporteurs in committees that are directly related to a given topic. Committees were manually matched with topics to achieve this. We control for whether or not an MEP serves in the EP leadership. Controls are also included for the total number of speeches made by an MEP and the percentage of MEP speeches that are available in English as these are liable to affect the observed MEP speech count. Finally, we also include dummy variables to control for an MEP's country of origin, EP party-group membership, and the topic on which they are speaking. All institutional and control variables were collected from the EP legislative observatory. 

\begin{figure}[!t]
	\centering
	\includegraphics[width=0.85\linewidth]{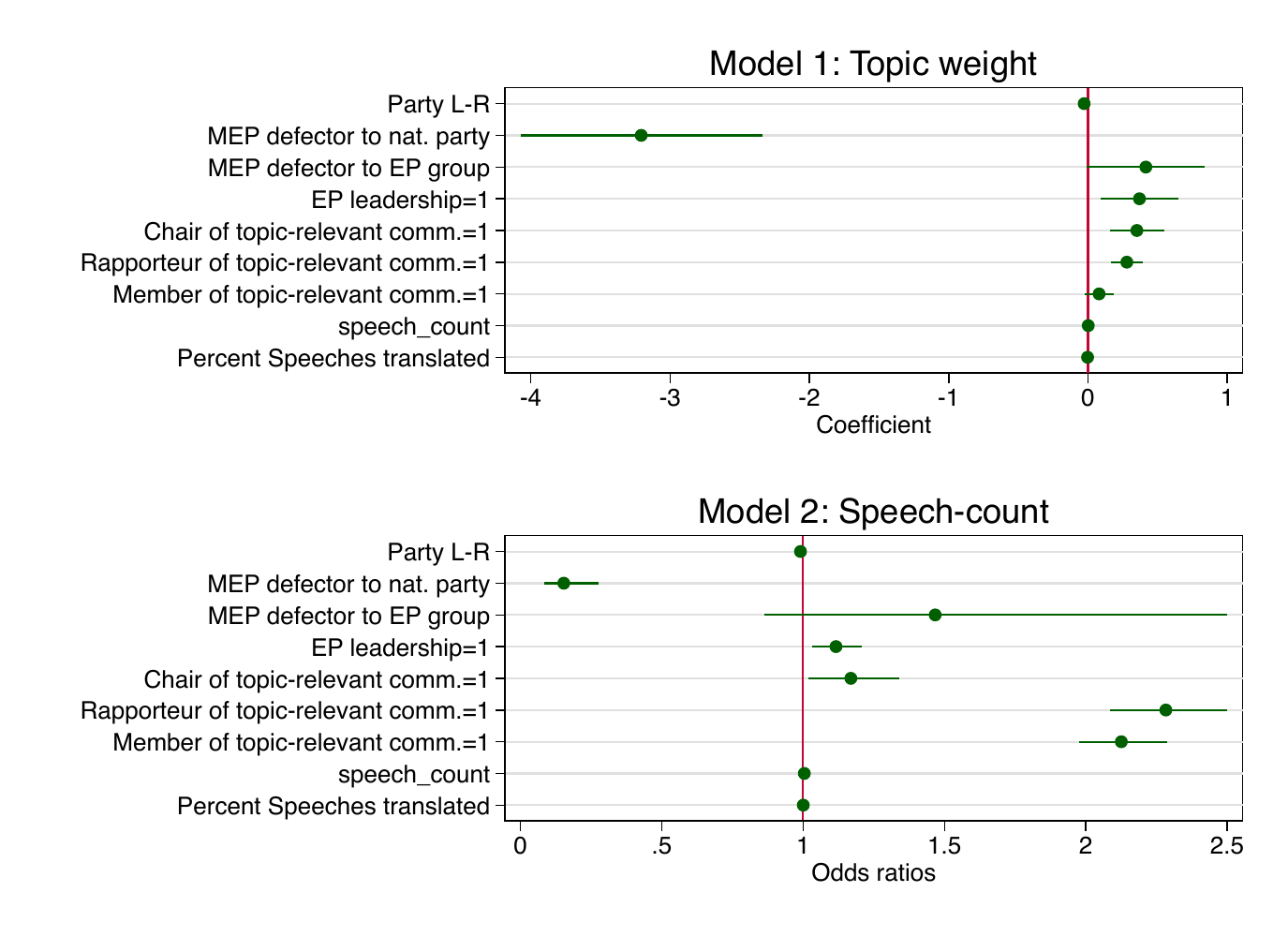}
	\caption{Plot of coefficients for regression models.}
	\label{fig:regfig}
\end{figure}

The regression results presented in \reffig{fig:regfig} provides further validation for our topic modeling approach. Model 1 provides strong evidence that voting behaviour and institutional position are the main drivers of MEP topic attention. MEPs that tend to defect from the European party group to vote with their national delegation tend to contribute to topics much less than those that are loyal to their party group, suggesting incentives to hide defection voting by avoiding making speeches when such votes are cast. MEPs with EP leadership roles are found to speak on topics more often, while the committee system is also a significant driver of MEP attention, with committee chairs and Rapporteurs making significantly more speeches on topics relevant to their official committee roles.

Turning to the Model 2, we see mostly similar substantive results, but this time we are presented with more directly interpretable odds ratios (exponentiated model coefficients) that capture the effects of our chosen independent variables on the odds of observing a speech relating to a given topic. The results suggest that the odds of MEPs who vote against their EP party groups in favor of their national party making a speech on a given topic are reduced by a factor of 0.86. The results also further reinforce our expectations that MEP positions within the EP committee system impact upon how much attention they pay to a particular topic. When an MEP holds a committee chair, Rapporteurship, or committee membership relevant to a particular topic, the odds that said MEP will make a speech on that topic increase by a factor of 1.169, 2.283, and 2.126 respectively. These results reinforce the idea that the committee system fundamentally shapes speech-making activities and the policy agenda of the EP plenary. The variable accounting for EP leadership positions is no longer significant in the count model, probably due to the fact that speeches belong to a single topic in this model rather than having multiple memberships, thus reducing the impact of EP leaders.

%% file: conclusion.tex
\section{Conclusions}
\label{sec:conc}

In this study, we propose a new two-layer NMF methodology for identifying topics in large political speech corpora over time, designed to identify both niche topics with specific and specialised vocabularies, and broader topics with more general vocabularies. Firstly, we demonstrate that topic modeling via NMF can lead to the identification of topics that are semantically more coherent in a corpus of political speeches, when compared with a probabilistic method such as LDA. Subsequently, we apply this method to a new corpus of all $\approx 210k$ English language plenary speeches from the EP between 1999--2014. In terms of providing substantive insight into EP politics, the topic modeling method allows us to unveil the political agenda of the EP, and the manner in which this agenda evolves over the time period considered. By considering three distinct case studies, we demonstrate the distinctions that can be drawn between the day-to-day political work of the EP in policy areas such as fisheries on the one hand, and the manner in which exogenous events such as economic crises and failed treaty referenda can give rise to new topics of discussion between MEPs on the other. With the EP agenda in hand, we explore the determinants of MEP attention to particular topics in the 7th sitting of the EP. We demonstrate how MEP voting behavior and institutional position affect whether or not they choose to contribute to an agenda topic. 

The insights provided by the dynamic topic modeling approach presented here demonstrate how these methods can uncover latent dynamics in MEP speech-making activities and supply new insights into how the EU functions as a political system. Much remains to be explored in terms of the patterns in political attention that emerge from our topic modeling approach. For instance, one would expect that political attention might well translate into influence over policy outcomes decided upon in the EP. Tracing influence to date has been difficult, as a macro-level picture of where and on what topics MEP attention lays has been unavailable. Linking political attention to political outcomes would help to unveil who gets what and when in European politics, which is a central concern for a political system often criticized for lacking democratic legitimacy.

Outside the European context, our method can be applied to any political situation in which policy agendas are captured in text form. Plenary debates in other political systems are a prime candidate for analysis, but legislative agendas, media agendas and other contexts where large corpora of text exist and are available digitally also lend themselves to analysis using our method.

%% file: appendix.tex
\newpage
\section*{Appendix A: Baseline Comparison}

Here we compare NMF and LDA\footnote{We use the LDA implementation provided by the MALLET toolkit: \url{http://mallet.cs.umass.edu}} when applied for topic modeling. Figures 1 and 2 show a full comparison of the coherence of the models generated by NMF and LDA on 60 time window datasets, for numbers of topics $k \in [10,50]$.
\begin{figure}[!h]	
	\centering
	\begin{subfigure}[t]{0.415\linewidth}
		\centering
		\includegraphics[width=\linewidth]{figures/coh-cv-k10.pdf}
		\caption{$C_{v}$ at $k=10$ topics}		
		\label{fig:cv-coh1}
	\end{subfigure}
	\hskip 0.5em
	\begin{subfigure}[t]{0.415\linewidth}
		\centering
		\includegraphics[width=\linewidth]{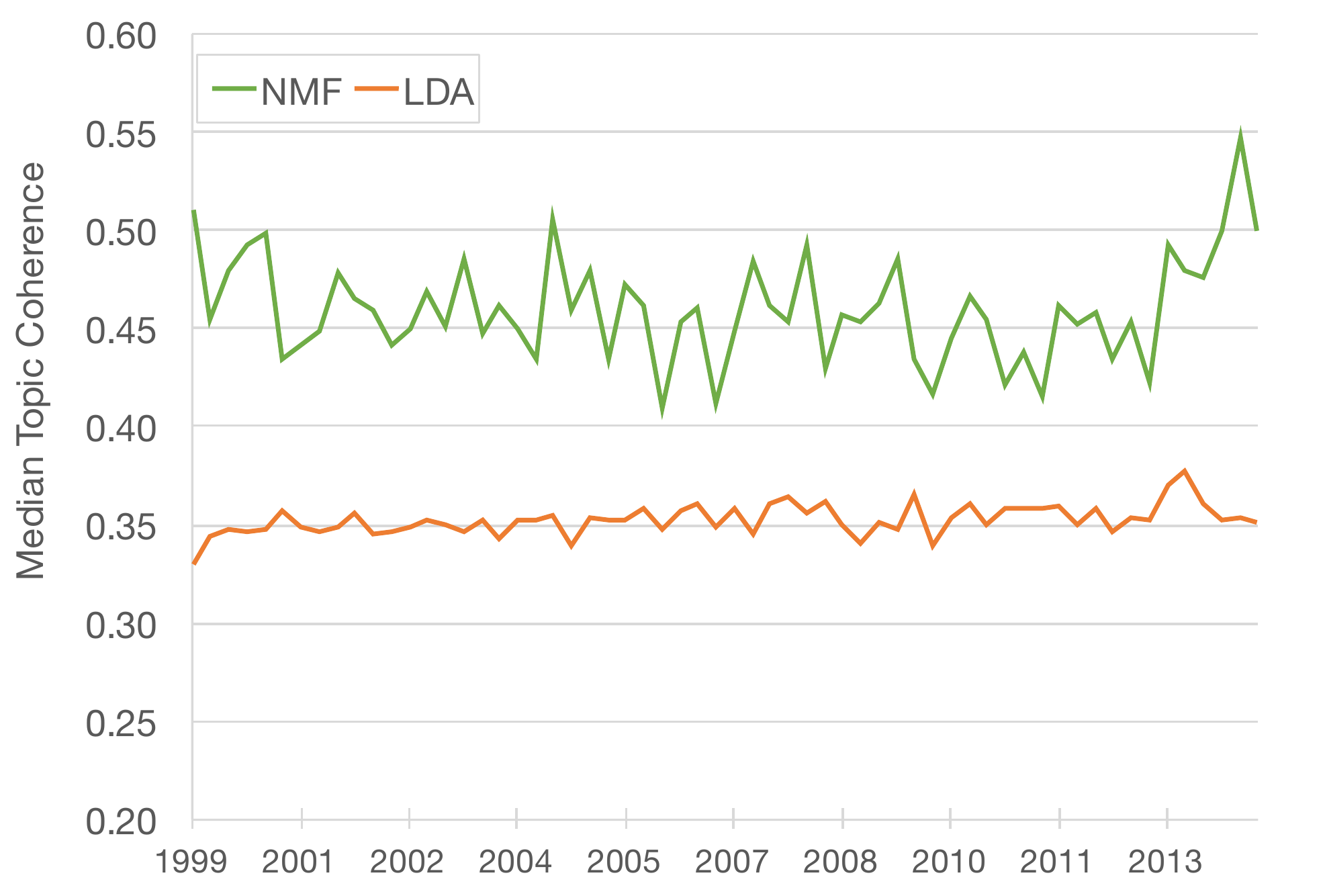}
		\caption{$C_{v}$ at $k=20$ topics}		
		\label{fig:cv-coh2}
	\end{subfigure}
	\vskip 0.5em
	\begin{subfigure}[t]{0.415\linewidth}
		\centering
		\includegraphics[width=\linewidth]{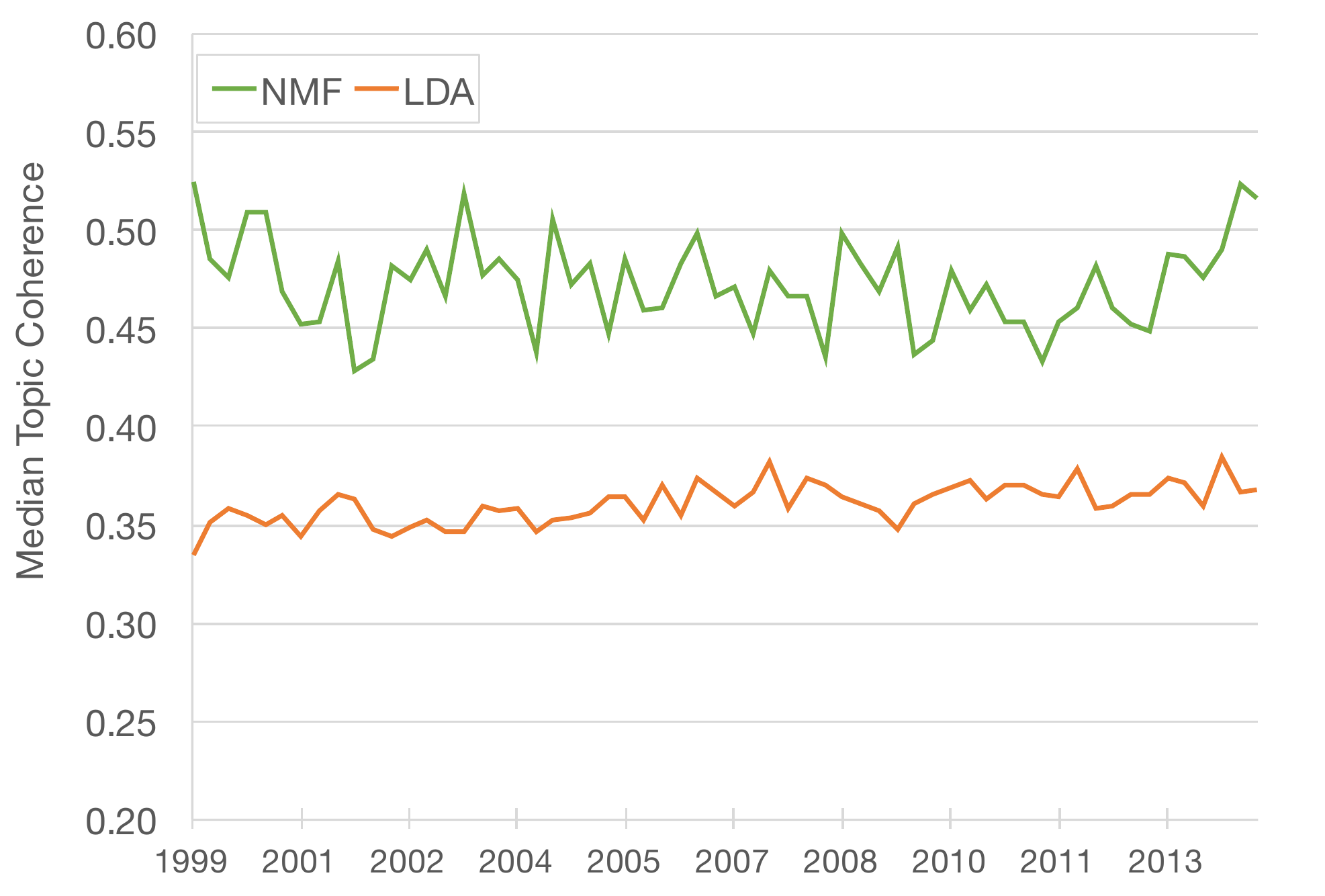}
		\caption{$C_{v}$  at $k=30$ topics}		
		\label{fig:cv-coh3}
	\end{subfigure}
	\hskip  0.5em
	\begin{subfigure}[t]{0.415\linewidth}
		\centering
		\includegraphics[width=\linewidth]{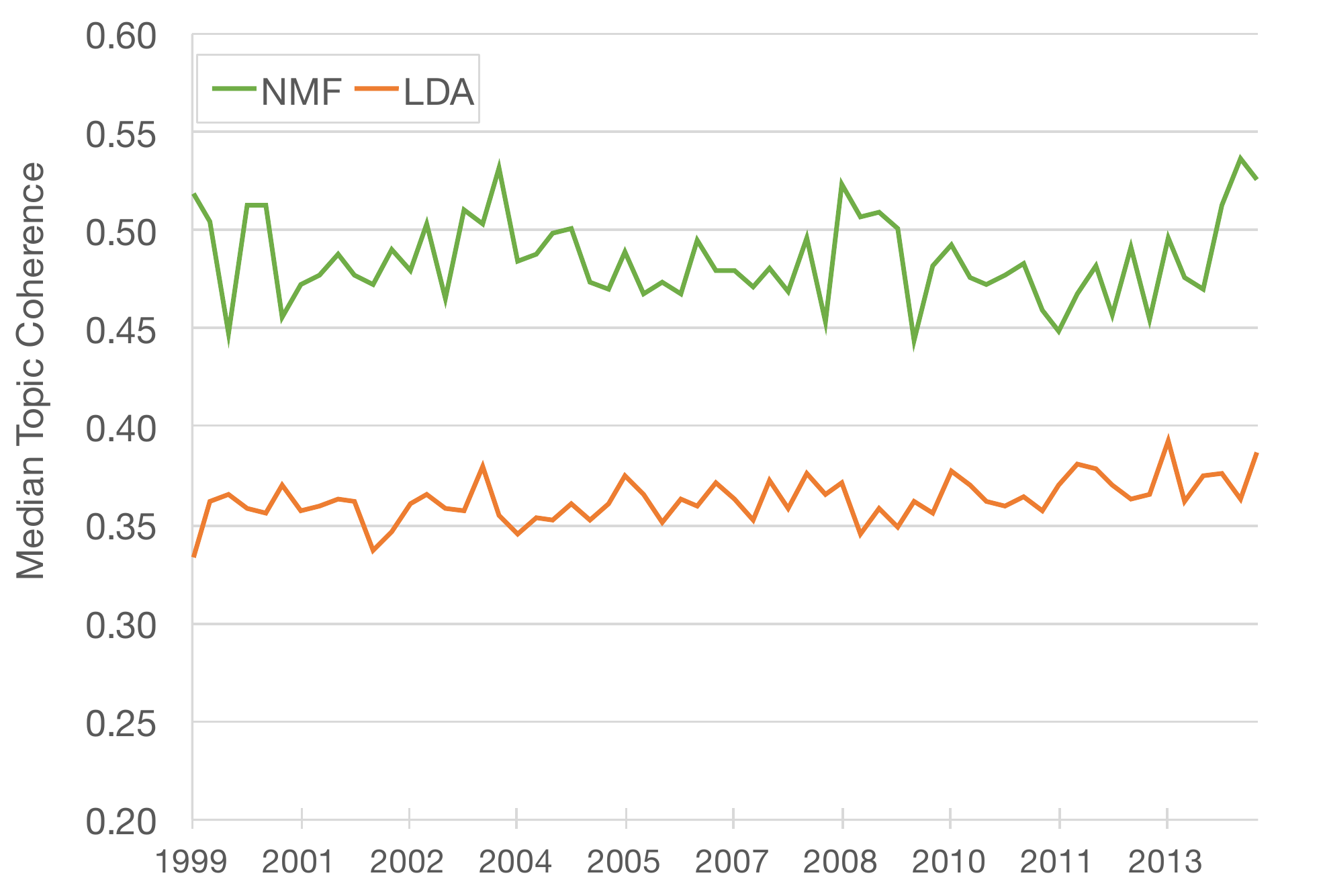}
		\caption{$C_{v}$  at $k=40$ topics}		
		\label{fig:cv-coh4}
	\end{subfigure}
	\vskip 0.5em
	\begin{subfigure}[t]{0.415\linewidth}
		\centering
		\includegraphics[width=\linewidth]{figures/coh-cv-k50.pdf}
		\caption{$C_{v}$ at $k=50$ topics}		
		\label{fig:cv-coh5}
	\end{subfigure}	
	\caption{Median $C_{v}$ topic coherence scores for all 60 time window datasets, for models produced by NMF and LDA with $k \in [10,50]$ topics.}
	\label{fig:cv-coh}
\end{figure}

\newpage
\begin{figure}[!h]	
	\centering
	\begin{subfigure}[t]{0.415\linewidth}
		\centering
		\includegraphics[width=\linewidth]{figures/coh-w2v-k10.pdf}
		\caption{TC-W2V at $k=10$ topics}		
		\label{fig:w2v-coh1}
	\end{subfigure}
	\hskip 0.5em
	\begin{subfigure}[t]{0.415\linewidth}
		\centering
		\includegraphics[width=\linewidth]{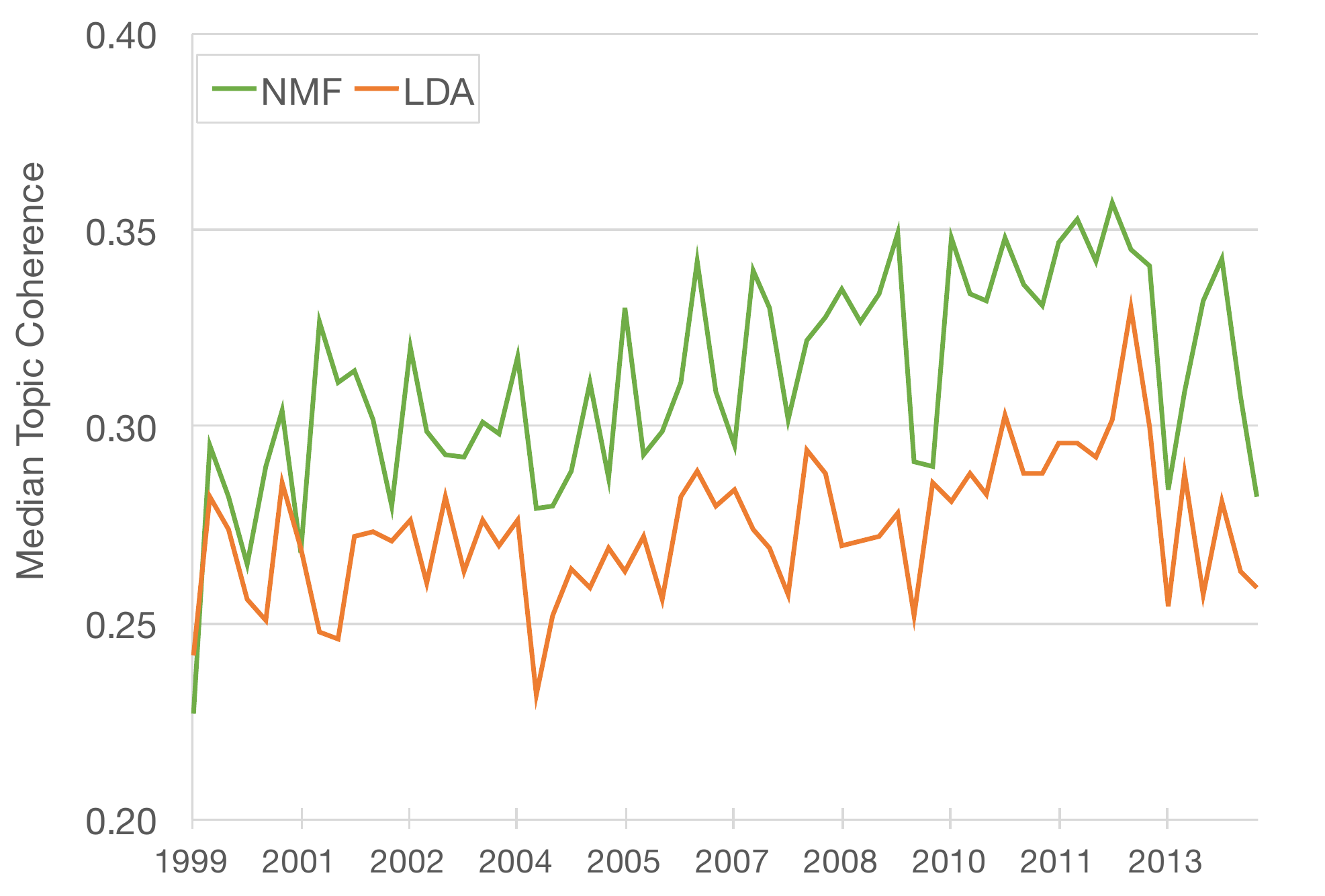}
		\caption{TC-W2V at $k=20$ topics}		
		\label{fig:w2v-coh2}
	\end{subfigure}
	\vskip 1em
	\begin{subfigure}[t]{0.415\linewidth}
		\centering
		\includegraphics[width=\linewidth]{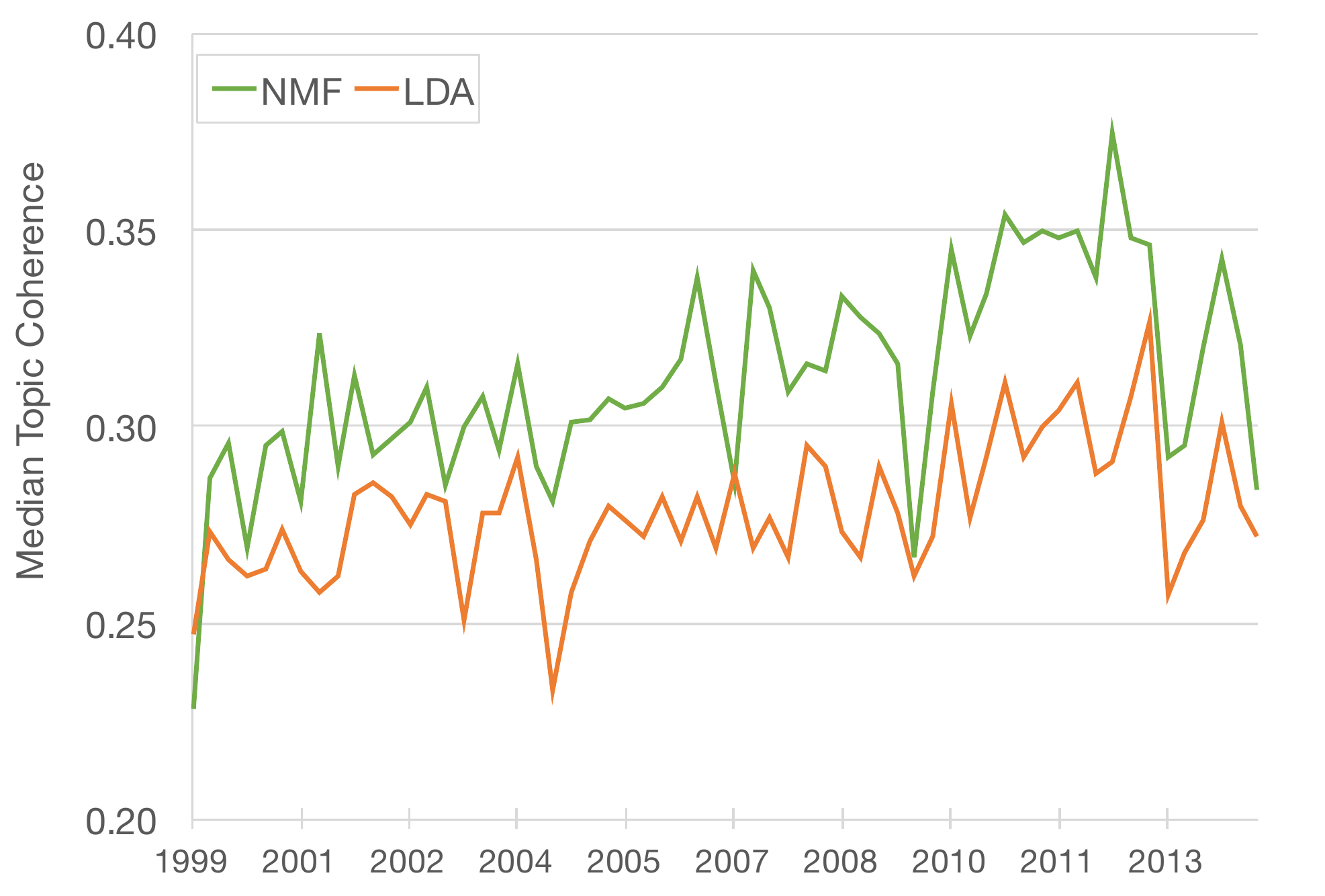}
		\caption{TC-W2V  at $k=30$ topics}		
		\label{fig:w2v-coh3}
	\end{subfigure}
	\hskip  0.5em
	\begin{subfigure}[t]{0.415\linewidth}
		\centering
		\includegraphics[width=\linewidth]{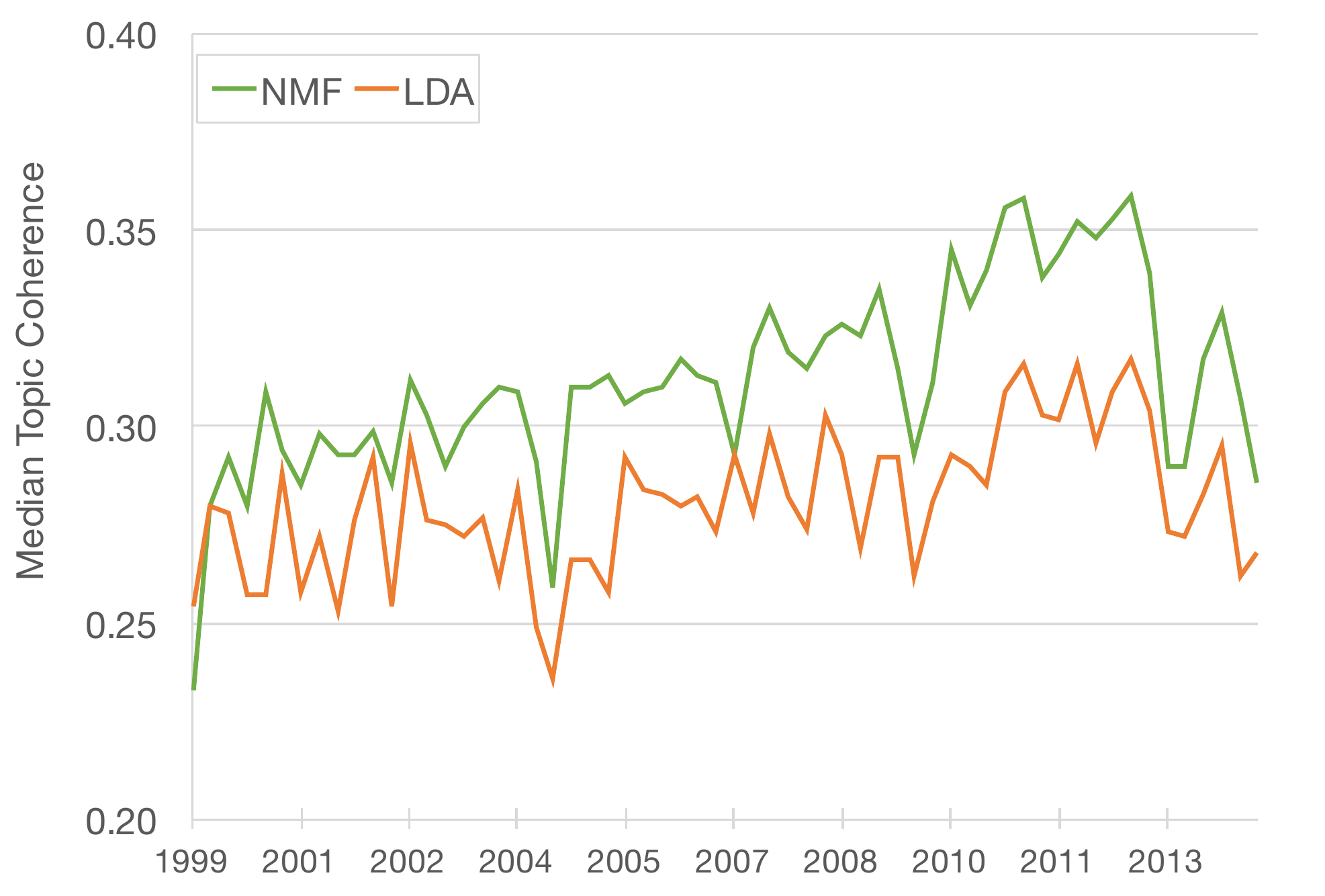}
		\caption{TC-W2V  at $k=40$ topics}		
		\label{fig:w2v-coh4}
	\end{subfigure}
	\vskip 1em
	\begin{subfigure}[t]{0.415\linewidth}
		\centering
		\includegraphics[width=\linewidth]{figures/coh-w2v-k50.pdf}
		\caption{TC-W2V at $k=50$ topics}		
		\label{fig:w2v-coh5}
	\end{subfigure}	
	\caption{Median TC-W2V topic coherence scores for all 60 time window datasets, for models produced by NMF and LDA with $k \in [10,50]$ topics.}
	\label{fig:w2v-coh}
\end{figure}

\section*{Appendix B: Topic Model Validation}
\label{sec:validation}

\subsection*{Intra-Topic Validity}

To examine the intra-topic semantic validity of the dynamic topics produced by our approach, we examined the distribution of TC-W2V coherence values for all dynamic topics, when evaluated in the \emph{word2vec} space built from the complete speech corpus. These coherence values correspond to the mean of the pairwise cosine similarities between the top-10 terms for each topic in the \emph{word2vec} space. As evidenced by the coherence values reported in \reftab{tab:topics}, the most coherent topics often correspond to core EU competencies. Unsurprisingly, broad administrative topics prove to be least coherent (\eg `Commission questions', `Council Presidency', `Plenary administration'). Overall the mean topic coherence score of $0.36$ is considerably higher than the lower bound for TC-W2V (\ie minimum value $= -1$), suggesting a high level of semantic validity.



\input{table_topics}

\subsection*{Inter-Topic Validity}

\begin{figure*}[!t]
    \centering
    \includegraphics[width=1\linewidth]{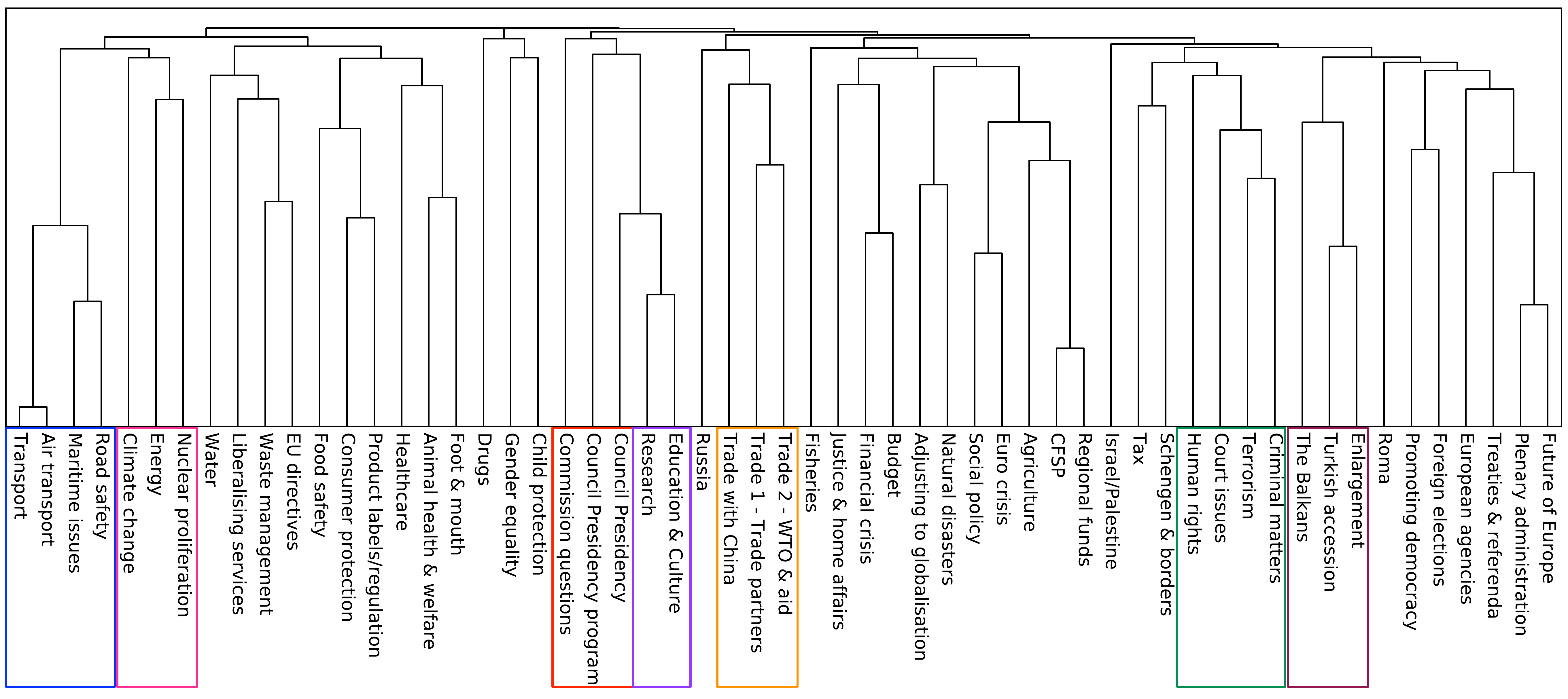}
    \vskip -0.7em
    \caption{Dendrogram for average linkage hierarchical agglomerative clustering of 57 dynamic topics.}
    \label{fig:tree}
\end{figure*}

To assess the inter-topic semantic validity of the results, we examine the extent to which any meaningful higher-level grouping exists among the 57 dynamic topics. To do this we apply average linkage agglomerative clustering to the topics. Using the approach described in \citet{greene08ensemble}, we re-cluster the row vectors from the second-layer NMF  factor $\m{H}$ using normalized Pearson correlation as a similarity metric. Here the vectors correspond the weights of each dynamic topic with respect to the 2,710 terms noted above. The dendrogram for the hierarchical clustering is shown in \reffig{fig:tree}. Following the interpretation provided in \citet{quinn2010analyze}, the lower the height at which any two topics are connected in the dendrogram, the more similar their term usage patterns in EP sessions. 

We observe a number of higher-level groupings of interest, which are highlighted in \reffig{fig:tree}. These includes groups specifically related to transport, (`Transport', `Air transport', `Maritime issues', `Road safety') energy (`Climate change', `Energy', `Nuclear proliferation'), animal health (`Drugs', `Foot \& mouth', `Animal health and welfare'), interactions with other institutions (`Council Presidency program', `Council Presidency', `Commission questions'), Education and research (`Education and Culture', `Research'), trade (`Trade with China', `Trade partnerships', `WTO \& aid'), and EU enlargement (`Enlargement', `Turkish accession', `The Balkans'). These hierarchical relationships between topics provide semantic validity for the model presented, where topics we would expect to be related are found to be correlated in the NMF factor $\m{H}$ (\ie they share similar terms). The presence of these higher-level associations between topics provide semantic validity for the results presented, where topics that one might expect to be related are found to be correlated with respect to rows in their NMF factor $\m{H}$ (\ie similar terms appear in the set of topic descriptors (words) that define them as topics).

\subsubsection*{External Validation}

The data analysis task performed in this paper is inherently unsupervised, in the sense that our corpus does not contain any annotated tags or labels indicating the nature of the content of speeches. Therefore, to assess the extent to which the dynamic topics identified correspond to EU policy areas, and thus provide evidence of construct validity, we compare the 57 dynamic topics to an existing taxonomy of subjects used by Europarl to classify legislative procedures. The taxonomy retrieved from the EP website has several different levels, ranging from broad top-level subjects (\eg `3 Community policies'), to highly-specific low-level subjects (\eg `3.10.06.05 Textile plants, cotton'). We compare our results to the second level of the taxonomy, containing 48 subjects (\eg `3.10 Agricultural policy and economies', `3.20 Transport policy in general'). For each subject code, we create a ``subject document'' consisting of the description of the subject and all lower-level subjects within that branch of the taxonomy. We then identify the most similar dynamic topic by comparing the top 10 terms for that topic with subject documents, based on cosine similarity. 

\input{table_subjects}

\begin{figure}[!t]
    \centering
    \includegraphics[width=0.7\linewidth]{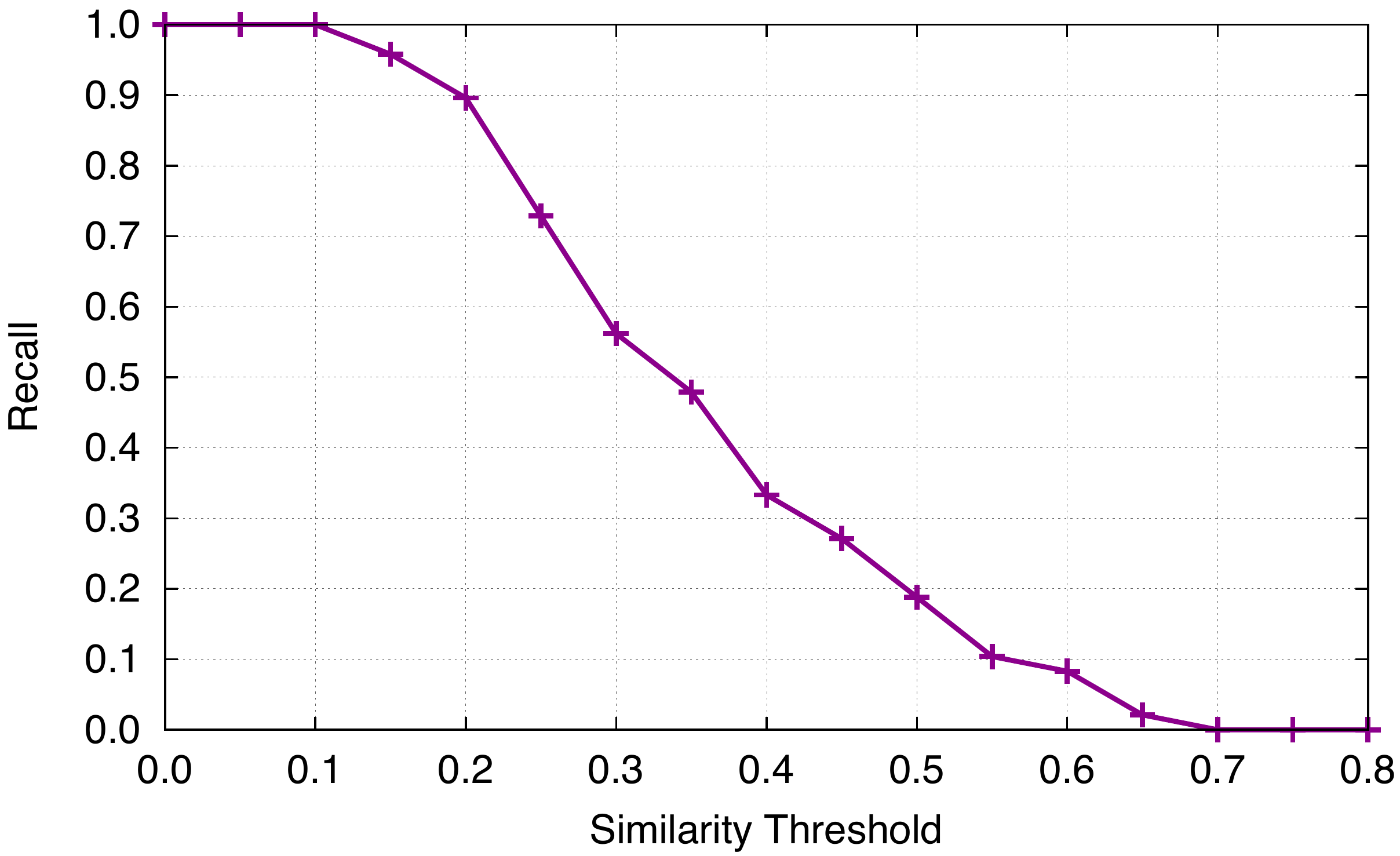}
	\caption{Recall plot for EP taxonomy subjects relative to dynamic topics, for increasing thresholds for cosine similarity.}
    \label{fig:subjects}
\end{figure}

\reftab{tab:subjects} shows the best matching subjects and topics identified using this approach. To give a couple of examples, the topic hand-coded as relating to `Tax' from our topic model was correctly matched with the Europarl subject code `2.70 Taxation' broadly defined at level-2 of the taxonomy, and with `2.70.01 Direct taxation' and `2.70.02 Indirect taxation' defined separately at level-3 of the taxonomy. When looking at the topic manually labeled as relating to `Drugs', cosine similarity matches this with the level-2 subject `4.20 Public health', which has a level-3 sub-category relating to `4.20.04 Pharmaceutical products and industry'. When taken in the context of the matches shown in \reftab{tab:subjects}, this indicates that our dynamic topics provide good coverage of the policy areas that might be expected to feature during EP debates, and thus increases our confidence in the construct validity of the model.

\newpage
\section*{Appendix C: Dynamic Comparison}
\label{sec:dtm}

As an additional comparison, we also examined the application of the probabilistic Dynamic Topic Modeling (DTM) algorithm proposed by \cite{blei06dynamic} to the corpus of parliamentary speeches. For the purpose of comparison, we apply both our proposed NMF-based approach and DTM for a fixed number of $k=50$ dynamic topics to the entire corpus. In the case of NMF, we generate the first layer of window topic models as described in Section 6.1 of the paper. In the case of DTM, we use the original C++ implementation\footnote{\url{http://www.cs.princeton.edu/~blei/topicmodeling.html}} and apply the algorithm using the default parameters recommended by the authors, using the same time window division as NMF.

\begin{figure}[!t]	
	\centering
	\begin{subfigure}[t]{0.495\linewidth}
		\centering
		\includegraphics[width=\linewidth]{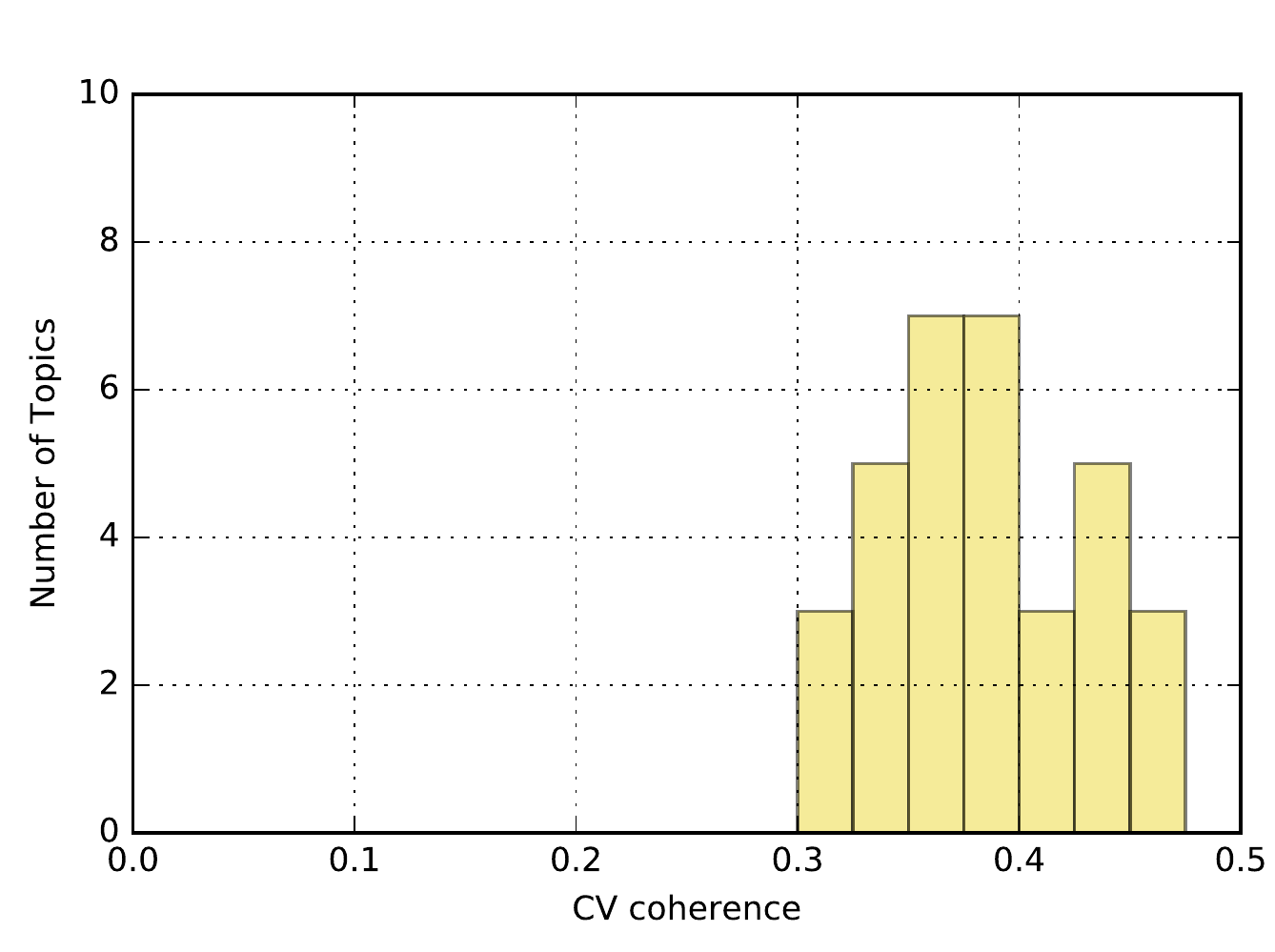}
		\caption{$C_{v}$ coherence for DTM}		
		\label{fig:dtm-cv-coh1}
	\end{subfigure}
	\hskip 0em
	\begin{subfigure}[t]{0.495\linewidth}
		\centering
		\includegraphics[width=\linewidth]{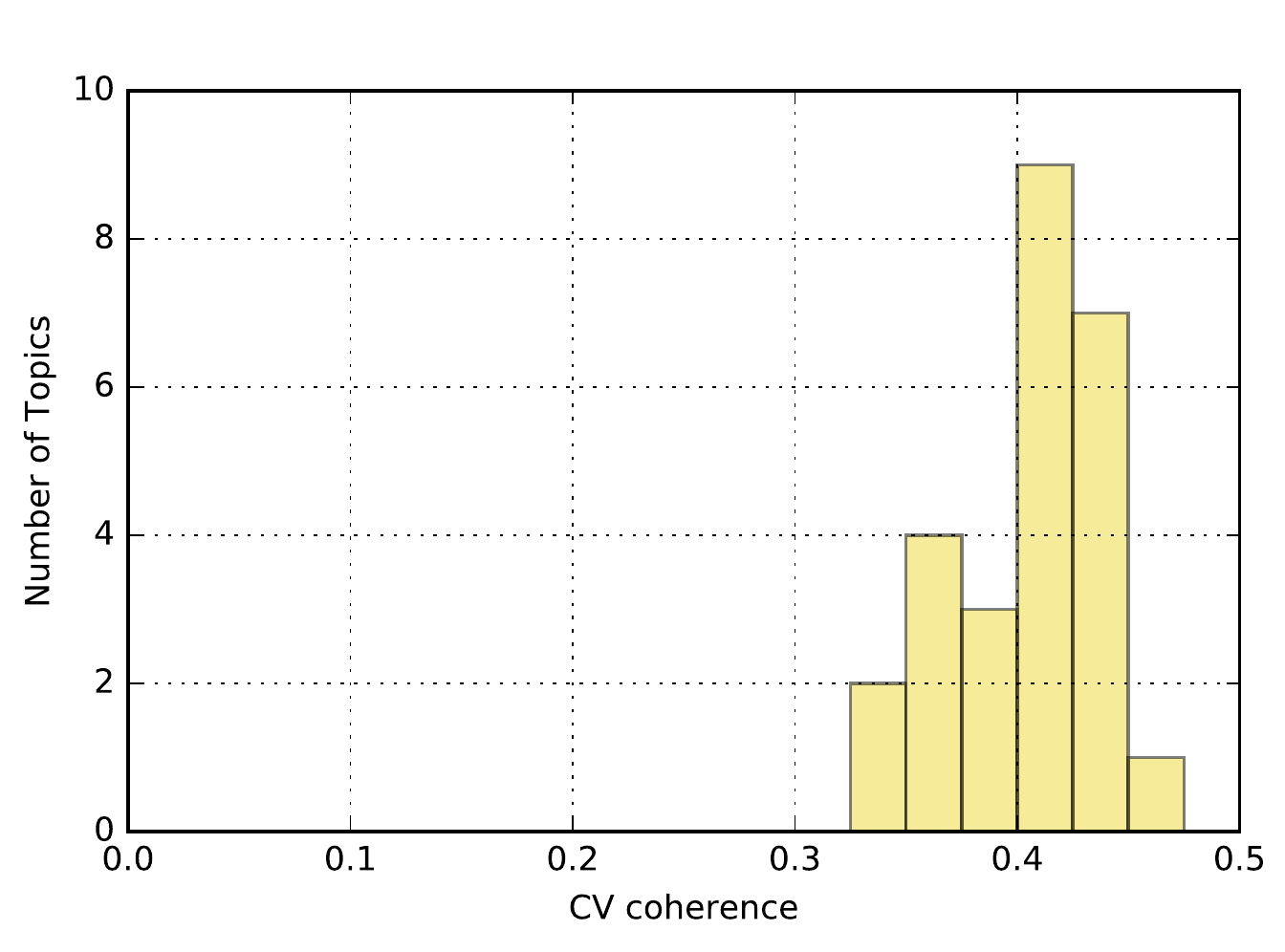}
		\caption{$C_{v}$ coherence for NMF}		
		\label{fig:dtm-cv-coh2}
	\end{subfigure}
	\begin{subfigure}[t]{0.495\linewidth}
		\centering
		\includegraphics[width=\linewidth]{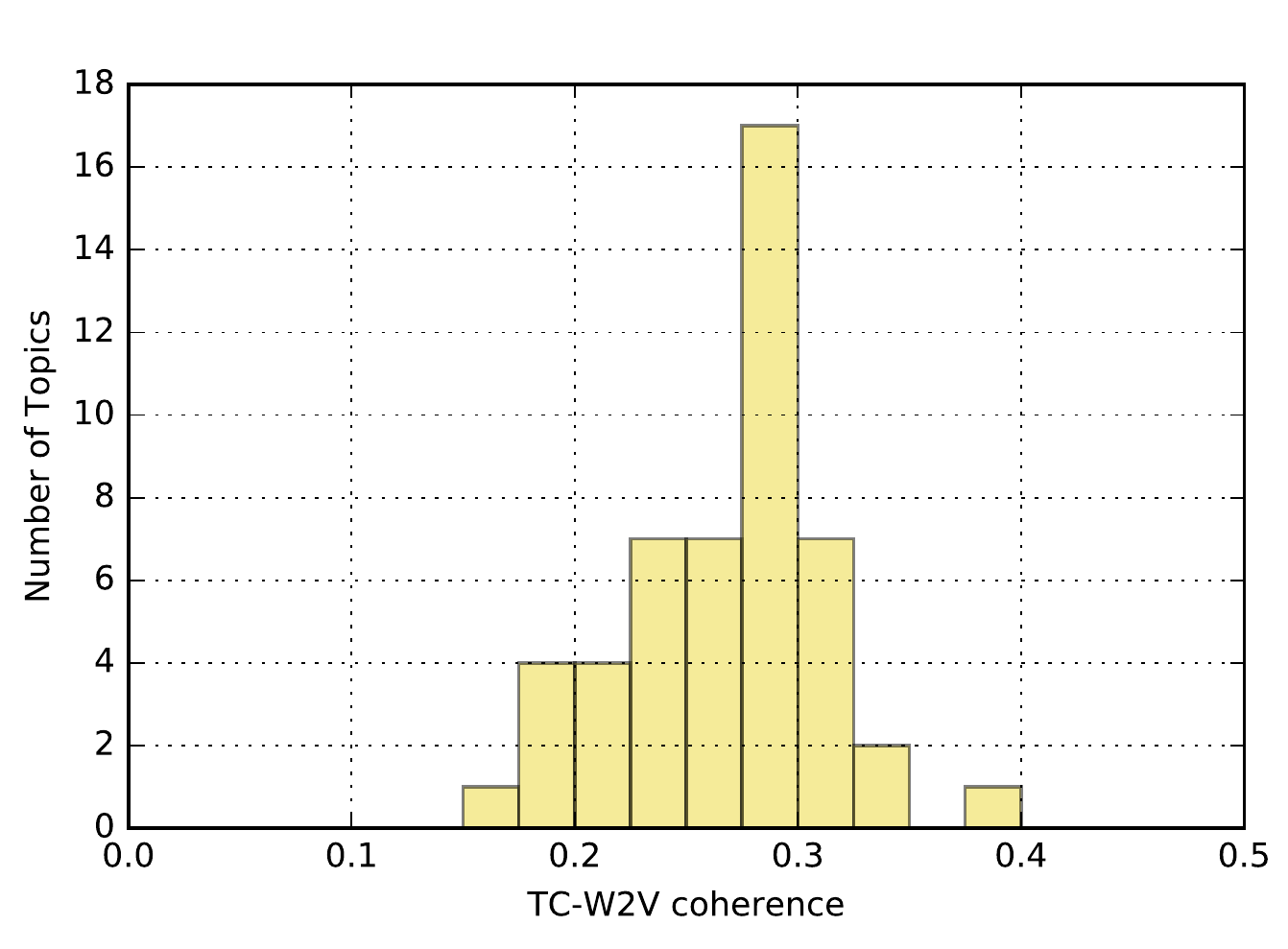}
		\caption{TC-W2V coherence for DTM}		
		\label{fig:dtm-w2v-coh1}
	\end{subfigure}
	\hskip 0em
	\begin{subfigure}[t]{0.495\linewidth}
		\centering
		\includegraphics[width=\linewidth]{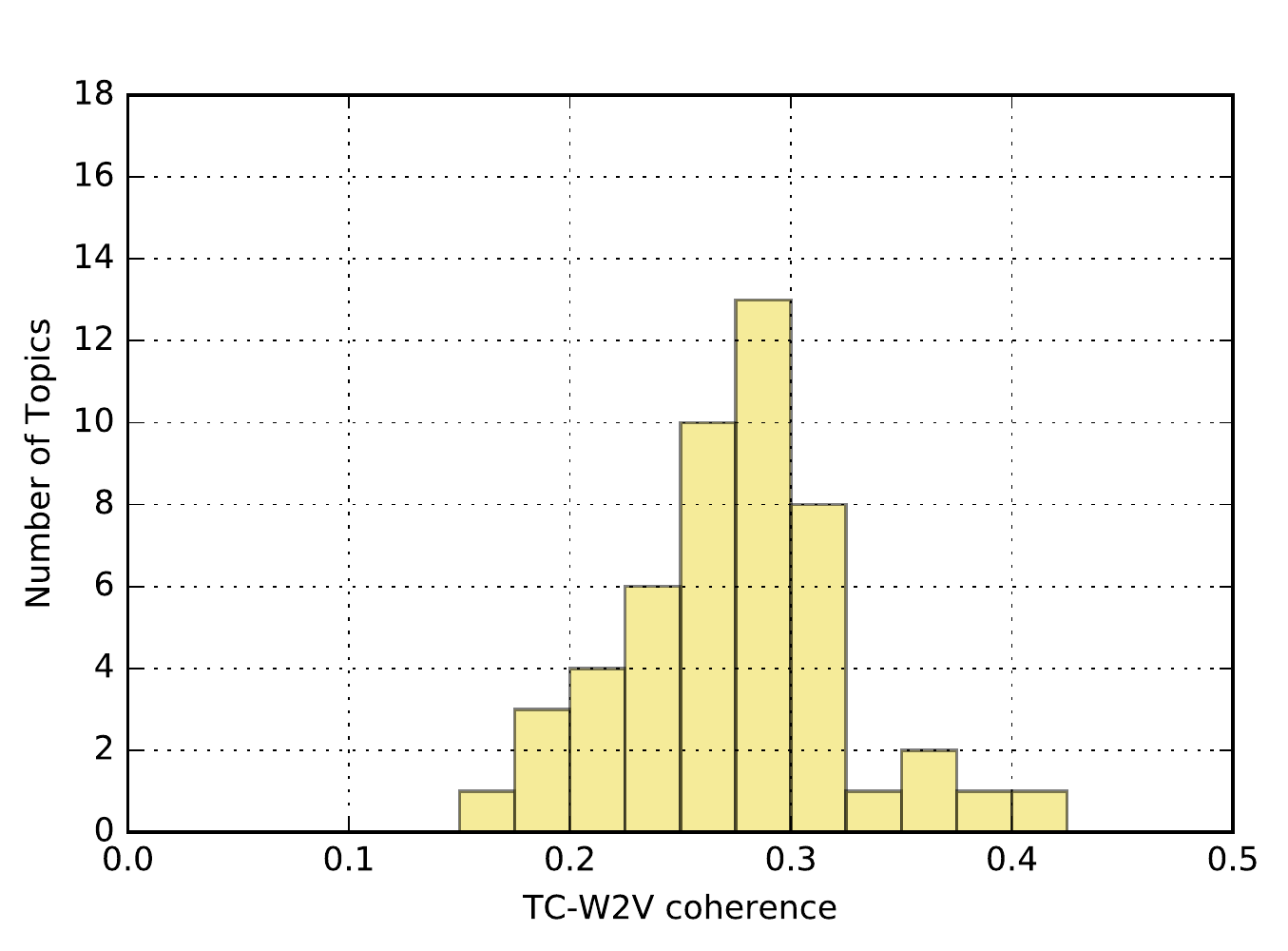}
		\caption{TC-W2V coherence for NMF}		
		\label{fig:dtm-w2v-coh2}
	\end{subfigure}
	\vskip -0.1em
	\caption{Distributions of coherence scores for $k=50$ dynamic topics, comparing the probabilistic and the NMF-based dynamic topic modeling methods.}
	\label{fig:dtm-w2v-coh}
\end{figure}

When we compare the overall results, the two approaches were in broad agreement, particularly in relation to the identification of dynamic topics relating to general policy areas, such as security, agriculture, transport, and fisheries. To quantitatively compare the outputs, we assessed the coherence of the dynamic topics using the TC-W2V and $C_{v}$ measures described in Section 5 of the paper, again using the top 10 terms to describe each topic. 
In the case of the $C_{v}$ topic coherence measure, the NMF approach had a higher median coherence of 0.458 versus 0.424 for DTM. The NMF-based approach also yielded a marginally higher median TC-W2V coherence of 0.277 versus 0.276. The distribution of values for all 50 dynamic topics are shown in \reffig{fig:dtm-w2v-coh}.

However, when we examine the actual window topics produced by each method, the results are quite different. Since the dynamic topics generated by DTM are built sequentially, the top terms reported at each time window are relatively stable. In contrast, with the NMF-based approach, each time window topic model is produced independently based only on the data present in that window. As a result, the top terms for each topic are far more indicative of the trends related to that topic at a given point in time. \reftab{tab:dtm} shows a representative example, corresponding to the dynamic topics related to climate change found by both methods, when broken down to their window topics across five quarterly time windows. We see that the top 10 terms for the NMF-based topics are far more diverse, reflecting the changing nature of discussion items around climate change in the European Parliament, such as the Cancun Agreements reached on at the 2010 United Nations Climate Change Conference in Mexico. 

\input{table_dtm}

To examine this difference quantitatively, for both topic modeling methods we look at the agreement between the top ranked terms consecutive pairs of window topics in each of the $k=50$ dynamic topics. We quantify the agreement between two term rankings using the Jaccard coefficient, which is the size of the intersection of the term sets divided by the size of their union. A score of 1 indicates that the term sets are identical (not considering rank order), while a score of 0 indicates that the sets share no terms in common. For each dynamic topic generated by the two methods, we calculate the mean agreement between the consecutive window topics form which it is composed. 

\reffig{fig:dtm-w2v-coh} shows the distribution of Jaccard agreement scores for the dynamic topics produced by both methods. We see a stark difference between the extent to which the terms associated with each topic change over time -- the overall mean Jaccard score across all dynamic topics for the NMF-based approach is 0.166, reflecting the fact that the top terms change frequently over time. In contrast, the overall mean score is 0.921 for the probabilistic approach indicates that the top terms often remain fixed and do not change frequently over time. Therefore, although the descriptors for the overall dynamic topics are relatively similar in terms of their coherence, when we wish to explore the time windows from which they are assembled, the NMF-based approach yields topics that more closely reflect the parliamentary discussions during each window, thereby supporting the interpretation of the topics.

\begin{figure}[!t]	
	\centering
	\begin{subfigure}[t]{0.495\linewidth}
		\centering
		\includegraphics[width=\linewidth]{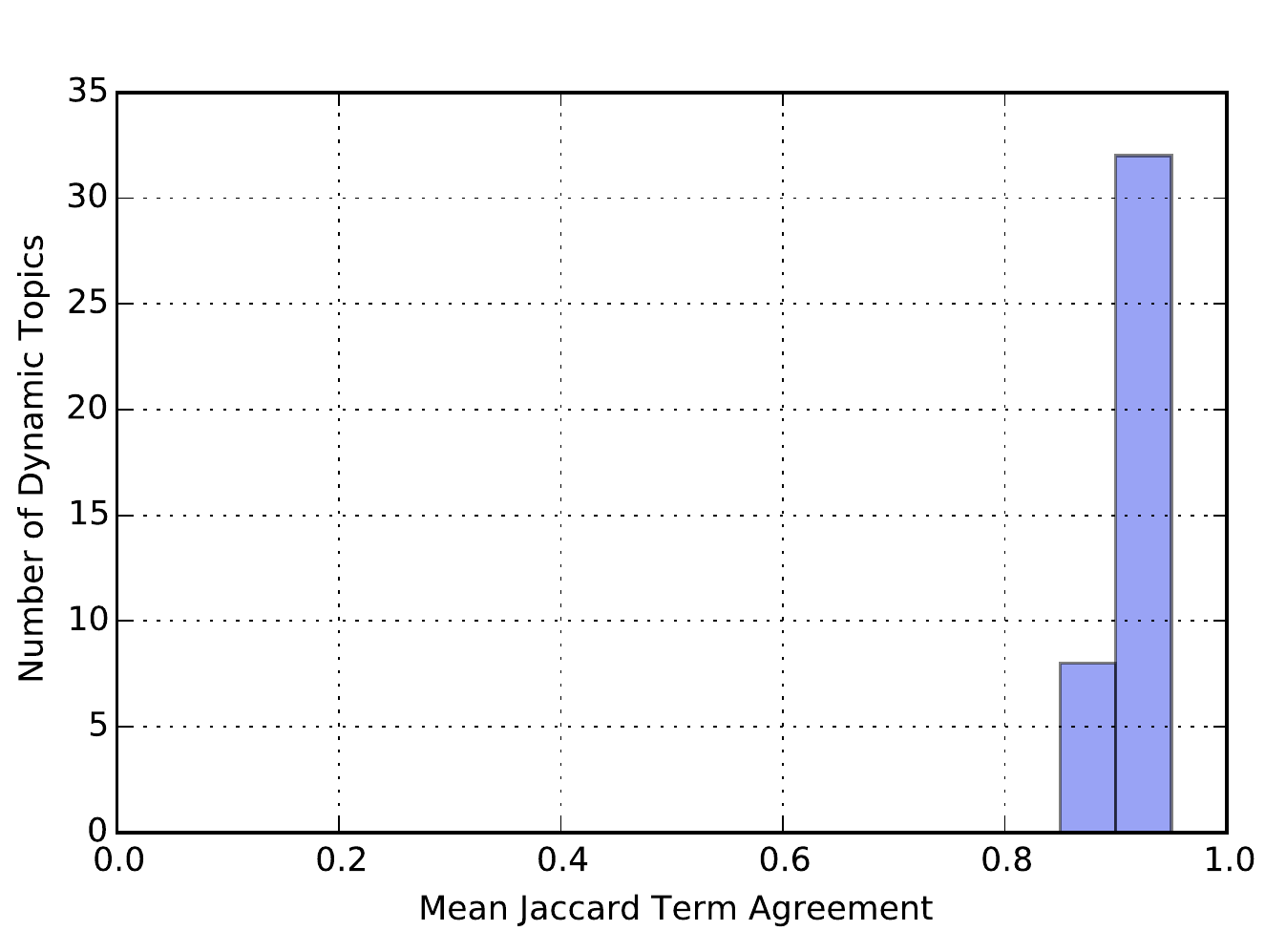}
		\caption{Jaccard Term Agreement for DTM}		
		\label{fig:dtm-jaccard1}
	\end{subfigure}
	\hskip 0em
	\begin{subfigure}[t]{0.495\linewidth}
		\centering
		\includegraphics[width=\linewidth]{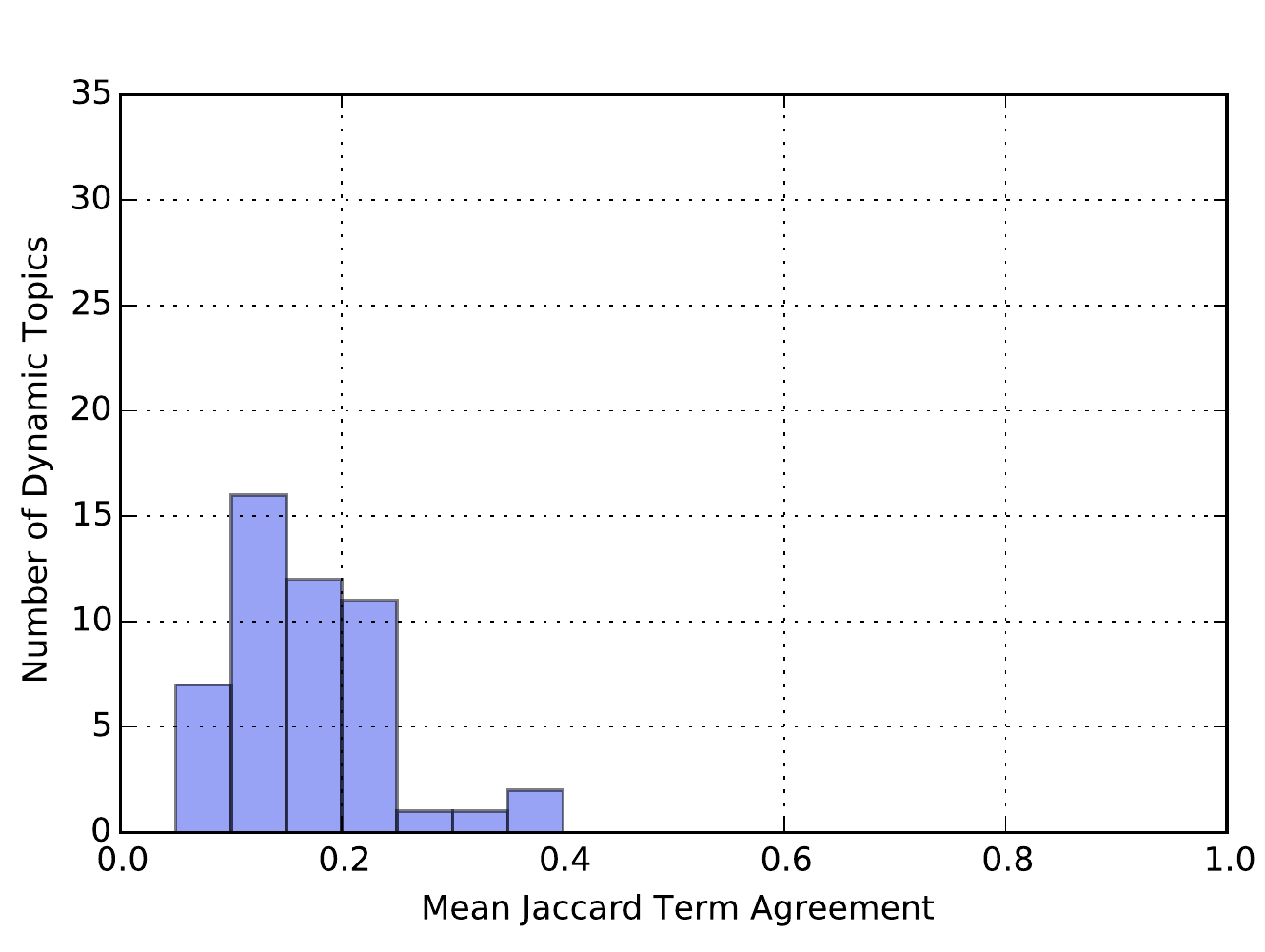}
		\caption{Jaccard Term Agreement for NMF}		
		\label{fig:dtm-jaccard2}
	\end{subfigure}
	\caption{Distributions of Jaccard term agreement scores $k=50$ dynamic topics, for the probabilistic and the NMF-based dynamic topic modeling methods.}
	\label{fig:dtm-jaccard}
\end{figure}

%% file: table_topics.tex
\begin{table*}[!t]
\centering
\scriptsize{
\begin{tabular}{|r|p{2.35cm}|p{7.25cm}|c|c|}
\hline
\textbf{Topic}& \textbf{Short Label}             & \textbf{Top 10 Terms}                                                                          & \textbf{Coh.} & \textbf{Freq.} \\\hline
13             & Transport                        & transport, railway, rail, passenger, road, network, freight, system, train, infrastructure              & 0.54          & 19             \\
42             & The Balkans                      & kosovo, serbia, balkan, resolution, bosnia, albania, iceland, herzegovina, macedonia, process           & 0.50          & 12             \\
33             & Air transport                    & air, passenger, transport, aviation, airport, traffic, airline, flight, sky, single                     & 0.48          & 10             \\
29             & Adjusting to globalisation       & fund, globalisation, egf, worker, adjustment, mobilisation, european, redundant, application, eur       & 0.47          & 15             \\
6              & Energy                           & energy, gas, renewable, efficiency, supply, source, electricity, market, target, project                & 0.47          & 36             \\
39             & Education \& culture             & programme, education, culture, language, cultural, youth, sport, learning, young, training              & 0.43          & 21             \\
8              & Fisheries                        & fishery, fishing, fish, stock, fisherman, fleet, sea, common, policy, measure                           & 0.43          & 34             \\
2              & Human rights                     & rights, human, fundamental, freedom, democracy, law, charter, resolution, union, violation              & 0.43          & 52             \\
45             & Maritime issues                  & port, sea, maritime, safety, ship, accident, oil, vessel, transport, inspection                         & 0.43          & 10             \\
21             & Healthcare                       & health, patient, environment, safety, public, care, healthcare, action, disease, mental                 & 0.42          & 18             \\
26             & Child protection                 & child, internet, pornography, sexual, school, exploitation, young, victim, education, crime             & 0.42          & 14             \\
56             & Road safety                      & road, safety, vehicle, transport, system, driver, accident, motor, noise, ecall                         & 0.41          & 12             \\
16             & Research                         & research, programme, innovation, framework, funding, industry, technology, development, cell, institute & 0.41          & 15             \\
15             & Turkish accession                & turkey, turkish, accession, progress, cyprus, negotiation, union, membership, croatia, macedonia        & 0.41          & 20             \\
35             & Tax                              & tax, vat, taxation, rate, system, fraud, states, evasion, car, transaction                              & 0.41          & 11             \\
32             & Trade - WTO \& aid            & trade, wto, world, development, developing, international, negotiation, aid, free, relation             & 0.39          & 19             \\
47             & Product labelling \& regulation & product, medicinal, medicine, tobacco, labelling, safety, consumer, regulation, organic, advertising    & 0.39          & 11             \\
11             & Trade - Trade partnerships     & agreement, partnership, morocco, trade, negotiation, data, cooperation, association, korea, fishery     & 0.39          & 18             \\
49             & Regional funds                   & policy, region, cohesion, development, regional, strategy, structural, fund, economic, area             & 0.39          & 22             \\
17             & CFSP                             & security, policy, defence, common, foreign, military, nato, immigration, aspect, european               & 0.39          & 19            \\\hline
\end{tabular}
}
\caption{List of top 20 dynamic topics, ranked by their TC-W2V topic coherence. For each dynamic topic, we report a manually-assigned short label, the top 10 terms, coherence, and frequency (\ie number of windows in which it appeared).}
\label{tab:topics}
\end{table*}

%% file: table_subjects.tex
\begin{table*}[!t]
\centering
\scriptsize{
\begin{tabular}{|l|p{6.45cm}|c|}
\hline
\textbf{Subject}                                     & \textbf{Matched Topic: Top 10 Terms}                                                                           & \textbf{Sim.} \\\hline
1.10\;\;Fundamental Rights In The Union                 & rights, human, fundamental, freedom, democracy, law, charter, resolution, union, violation                    & 0.66          \\
4.40\;\;Education, Vocational Training \& Youth                 & programme, education, culture, language, cultural, youth, sport, learning, young, training                    & 0.63          \\
5.20\;\;Monetary Union                                           & euro, economic, growth, stability, pact, bank, policy, monetary, economy, ecb                                 & 0.62          \\
4.70\;\;Regional Policy                                          & policy, region, cohesion, development, regional, strategy, structural, fund, economic, area                   & 0.62          \\
3.50\;\;Research \& Technological Development               & research, programme, innovation, framework, funding, industry, technology, development, cell, institute       & 0.57          \\
3.60\;\;Energy Policy                                            & energy, gas, renewable, efficiency, supply, source, electricity, market, target, project                      & 0.53          \\
6.10\;\;Common Foreign \& Security Policy               & security, policy, defence, common, foreign, military, nato, immigration, aspect, european                     & 0.52          \\
3.20\;\;Transport Policy in General                              & transport, railway, rail, passenger, road, network, freight, system, train, infrastructure                    & 0.51          \\
4.60\;\;Consumers' Protection in General                         & product, medicinal, medicine, tobacco, labelling, safety, consumer, regulation, organic, advertising          & 0.50          \\
3.70\;\;Environmental Policy                                     & waste, recycling, directive, packaging, management, environment, electronic, fuel, environmental, radioactive & 0.50          \\\hline
\end{tabular}
}
\caption{Top 10 legislative procedure subjects with corresponding matching dynamic topics, ranked by cosine similarity of the match.}
\label{tab:subjects}
\end{table*}

%% file: table_dtm.tex
\begin{table*}[!t]
\centering
\scriptsize{
\begin{tabular}{|l|p{5.75cm}|p{5.75cm}|}
\hline
\textbf{Window}                                     & \textbf{NMF Window Topic}                                                                           & \textbf{DTM Window Topic}      \\\hline
2008-Q4& energy, climate, emission, package, change, renewable, target, industry, carbon, gas & energy, climate, change, gas, european, emission, package, supply, efficiency, renewable\\
2009-Q1& climate, change, future, emission, integrated, water, policy, target, industrial, global	& energy, climate, change, gas, european, emission, efficiency, supply, package, renewable\\
2009-Q4& climate, change, copenhagen, developing, emission, conference, summit, agreement, global, energy	& energy, climate, change, copenhagen, european, gas, emission, efficiency, supply, carbon\\
2010-Q1& climate, copenhagen, change, summit, emission, international, mexico, conference, global, world & energy, climate, change, european, copenhagen, gas, emission, efficiency, supply, carbon\\
2010-Q4& climate, trade, change, cancun, conference, international, agreement, emission, environmental, global & energy, climate, change, european, gas, efficiency, emission, supply, target, source\\\hline
\end{tabular}
}
\caption{Example of window topics associated with a dynamic topic related to climate change, produced by both the NMF-based approach and DTM on the same time window datasets.}
\label{tab:dtm}
\end{table*}

%% file: exploring-arxiv.bbl
\begin{thebibliography}{}

\bibitem[\protect\citeauthoryear{Alexandrova, Carammia, and
  Timmermans}{Alexandrova et~al.}{2012}]{alexandrova2012policy}
Alexandrova, P., M.~Carammia, and A.~Timmermans (2012).
\newblock Policy punctuations and issue diversity on the european council
  agenda.
\newblock {\em Policy Studies Journal\/}~{\em 40\/}(1), 69--88.

\bibitem[\protect\citeauthoryear{Baumgartner, Breunig, Green-Pedersen, Jones,
  Mortensen, Nuytemans, and Walgrave}{Baumgartner
  et~al.}{2009}]{baumgartner2009punctuated}
Baumgartner, F.~R., C.~Breunig, C.~Green-Pedersen, B.~D. Jones, P.~B.
  Mortensen, M.~Nuytemans, and S.~Walgrave (2009).
\newblock Punctuated equilibrium in comparative perspective.
\newblock {\em American Journal of Political Science\/}~{\em 53\/}(3),
  603--620.

\bibitem[\protect\citeauthoryear{Baumgartner and Jones}{Baumgartner and
  Jones}{1993}]{baumgartner1993agendas}
Baumgartner, F.~R. and B.~D. Jones (1993).
\newblock {\em Agendas and Instability in American Politics}.
\newblock University of Chicago Press.

\bibitem[\protect\citeauthoryear{Blei and Lafferty}{Blei and
  Lafferty}{2006}]{blei06dynamic}
Blei, D.~M. and J.~D. Lafferty (2006).
\newblock Dynamic topic models.
\newblock In {\em {Proc. 23rd International Conference on Machine Learning}},
  pp.\  113--120.

\bibitem[\protect\citeauthoryear{Blei, Ng, and Jordan}{Blei
  et~al.}{2003}]{blei03lda}
Blei, D.~M., A.~Y. Ng, and M.~I. Jordan (2003).
\newblock Latent dirichlet allocation.
\newblock {\em Journal of Machine Learning Research\/}~{\em 3}, 993--1022.

\bibitem[\protect\citeauthoryear{Boutsidis and Gallopoulos}{Boutsidis and
  Gallopoulos}{2008}]{bout08headstart}
Boutsidis, C. and E.~Gallopoulos (2008).
\newblock {SVD based initialization: A head start for non-negative matrix
  factorization}.
\newblock {\em Pattern Recognition\/}.

\bibitem[\protect\citeauthoryear{Bowler and Farrell}{Bowler and
  Farrell}{1995}]{bowler1995organizing}
Bowler, S. and D.~M. Farrell (1995).
\newblock The organizing of the european parliament: Committees, specialization
  and co-ordination.
\newblock {\em British J. Political Science\/}~{\em 25\/}(02), 219--243.

\bibitem[\protect\citeauthoryear{Cameron and Trivedi}{Cameron and
  Trivedi}{2013}]{cameron2013regression}
Cameron, A.~C. and P.~K. Trivedi (2013).
\newblock {\em Regression analysis of count data}, Volume~53.
\newblock Cambridge university press.

\bibitem[\protect\citeauthoryear{Chang, Boyd-Graber, Gerrish, Wang, and
  Blei}{Chang et~al.}{2009}]{chang09tea}
Chang, J., J.~Boyd-Graber, S.~Gerrish, C.~Wang, and D.~M. Blei (2009).
\newblock {Reading Tea Leaves: How Humans Interpret Topic Models}.
\newblock In {\em NIPS}, pp.\  288--296.

\bibitem[\protect\citeauthoryear{Deerwester, Dumais, Landauer, Furnas, and
  Harshman}{Deerwester et~al.}{1990}]{deerwester90lsi}
Deerwester, S.~C., S.~T. Dumais, T.~K. Landauer, G.~W. Furnas, and R.~A.
  Harshman (1990).
\newblock Indexing by latent semantic analysis.
\newblock {\em Journal of the American Society of Information Science\/}~{\em
  41\/}(6), 391--407.

\bibitem[\protect\citeauthoryear{Dowding, Hindmoor, and Martin}{Dowding
  et~al.}{2015}]{dowding2015comparative}
Dowding, K., A.~Hindmoor, and A.~Martin (2015).
\newblock The comparative policy agendas project: theory, measurement and
  findings.
\newblock {\em Journal of Public Policy\/}, 1--23.

\bibitem[\protect\citeauthoryear{Downs}{Downs}{1972}]{downs1972up}
Downs, A. (1972).
\newblock Up and down with ecology-the issue-attention cycle.
\newblock {\em The public interest\/}~(28), 38.

\bibitem[\protect\citeauthoryear{Greene, Cagney, Krogan, and Cunningham}{Greene
  et~al.}{2008}]{greene08ensemble}
Greene, D., G.~Cagney, N.~Krogan, and P.~Cunningham (2008).
\newblock {Ensemble Non-negative Matrix Factorization Methods for Clustering
  Protein-Protein Interactions}.
\newblock {\em Bioinformatics\/}~{\em 24\/}(15), 1722--1728.

\bibitem[\protect\citeauthoryear{Grimmer}{Grimmer}{2010}]{grimmer10bayesian}
Grimmer, J. (2010).
\newblock A bayesian hierarchical topic model for political texts: Measuring
  expressed agendas in senate press releases.
\newblock {\em Political Analysis\/}~{\em 18\/}(1), 1--35.

\bibitem[\protect\citeauthoryear{Hix, Noury, and Roland}{Hix
  et~al.}{2006}]{hix2006dimensions}
Hix, S., A.~Noury, and G.~Roland (2006).
\newblock Dimensions of politics in the european parliament.
\newblock {\em American J. Political Science\/}~{\em 50\/}(2), 494--520.

\bibitem[\protect\citeauthoryear{{Hix, Simon}, Noury, and Roland}{{Hix, Simon}
  et~al.}{2007}]{hix2007democratic}
{Hix, Simon}, A.~Noury, and G.~Roland (2007).
\newblock {\em {Democratic politics in the European Parliament}}.
\newblock Cambridge University Press.

\bibitem[\protect\citeauthoryear{Jennings, Bevan, Timmermans, Breeman, Brouard,
  Chaqu{\'e}s-Bonafont, Green-Pedersen, John, Mortensen, and Palau}{Jennings
  et~al.}{2011}]{jennings2011effects}
Jennings, W., S.~Bevan, A.~Timmermans, G.~Breeman, S.~Brouard,
  L.~Chaqu{\'e}s-Bonafont, C.~Green-Pedersen, P.~John, P.~B. Mortensen, and
  A.~M. Palau (2011).
\newblock Effects of the core functions of government on the diversity of
  executive agendas.
\newblock {\em Comparative Political Studies\/}, 0010414011405165.

\bibitem[\protect\citeauthoryear{John and Bevan}{John and
  Bevan}{2012}]{john2012policy}
John, P. and S.~Bevan (2012).
\newblock What are policy punctuations? large changes in the legislative agenda
  of the uk government, 1911--2008.
\newblock {\em Policy Studies Journal\/}~{\em 40\/}(1), 89--108.

\bibitem[\protect\citeauthoryear{Jones}{Jones}{1994}]{jones1994reconceiving}
Jones, B.~D. (1994).
\newblock {\em Reconceiving decision-making in democratic politics: Attention,
  choice, and public policy}.
\newblock University of Chicago Press.

\bibitem[\protect\citeauthoryear{Jones and Baumgartner}{Jones and
  Baumgartner}{2005}]{jones2005politics}
Jones, B.~D. and F.~R. Baumgartner (2005).
\newblock {\em The politics of attention: How government prioritizes problems}.
\newblock University of Chicago Press.

\bibitem[\protect\citeauthoryear{Jones and Baumgartner}{Jones and
  Baumgartner}{2012}]{jones2012there}
Jones, B.~D. and F.~R. Baumgartner (2012).
\newblock From there to here: Punctuated equilibrium to the general punctuation
  thesis to a theory of government information processing.
\newblock {\em Policy Studies Journal\/}~{\em 40\/}(1), 1--20.

\bibitem[\protect\citeauthoryear{Lee and Seung}{Lee and Seung}{1999}]{lee99nmf}
Lee, D.~D. and H.~S. Seung (1999).
\newblock Learning the parts of objects by non-negative matrix factorization.
\newblock {\em Nature\/}~{\em 401}, 788--91.

\bibitem[\protect\citeauthoryear{McCallum}{McCallum}{2002}]{mccallum02mallet}
McCallum, A.~K. (2002).
\newblock {Mallet: A machine learning for language toolkit}.

\bibitem[\protect\citeauthoryear{Mikolov, Chen, Corrado, and Dean}{Mikolov
  et~al.}{2013}]{mikolovEfficient}
Mikolov, T., K.~Chen, G.~Corrado, and J.~Dean (2013).
\newblock Efficient estimation of word representations in vector space.
\newblock {\em CoRR\/}~{\em abs/1301.3781}.

\bibitem[\protect\citeauthoryear{O'Callaghan, Greene, Carthy, and
  Cunningham}{O'Callaghan et~al.}{2015}]{ocallaghan15eswa}
O'Callaghan, D., D.~Greene, J.~Carthy, and P.~Cunningham (2015).
\newblock An analysis of the coherence of descriptors in topic modeling.
\newblock {\em Expert Systems with Applications (ESWA)\/}~{\em 42\/}(13),
  5645--5657.

\bibitem[\protect\citeauthoryear{Proksch and Slapin}{Proksch and
  Slapin}{2010}]{proksch2010position}
Proksch, S.-O. and J.~B. Slapin (2010).
\newblock {Position taking in European Parliament speeches}.
\newblock {\em British J. Political Science\/}~{\em 40\/}(03), 587--611.

\bibitem[\protect\citeauthoryear{Proksch and Slapin}{Proksch and
  Slapin}{2014}]{proksch2014politics}
Proksch, S.-O. and J.~B. Slapin (2014).
\newblock {\em The politics of parliamentary debate: Parties, rebels and
  representation}.
\newblock Cambridge University Press.

\bibitem[\protect\citeauthoryear{Quinn, Monroe, Colaresi, Crespin, and
  Radev}{Quinn et~al.}{2010}]{quinn2010analyze}
Quinn, K.~M., B.~L. Monroe, M.~Colaresi, M.~H. Crespin, and D.~R. Radev (2010).
\newblock How to analyze political attention with minimal assumptions and
  costs.
\newblock {\em American J. Political Science\/}~{\em 54\/}(1), 209--228.

\bibitem[\protect\citeauthoryear{Roberts, Stewart, Tingley, Lucas, Leder-Luis,
  Gadarian, Albertson, and Rand}{Roberts et~al.}{2014}]{roberts2014structural}
Roberts, M.~E., B.~M. Stewart, D.~Tingley, C.~Lucas, J.~Leder-Luis, S.~K.
  Gadarian, B.~Albertson, and D.~G. Rand (2014).
\newblock {Structural Topic Models for Open-Ended Survey Responses}.
\newblock {\em American J. Political Science\/}~{\em 58\/}(4), 1064--1082.

\bibitem[\protect\citeauthoryear{R{\"o}der, Both, and Hinneburg}{R{\"o}der
  et~al.}{2015}]{roder15palmetto}
R{\"o}der, M., A.~Both, and A.~Hinneburg (2015).
\newblock Exploring the space of topic coherence measures.
\newblock In {\em Proc. 8th ACM international conference on Web search and data
  mining}, pp.\  399--408. ACM.

\bibitem[\protect\citeauthoryear{Scully, Hix, and Farrell}{Scully
  et~al.}{2012}]{scully2012national}
Scully, R., S.~Hix, and D.~M. Farrell (2012).
\newblock {National or European Parliamentarians? Evidence from a New Survey of
  the Members of the European Parliament}.
\newblock {\em JCMS: J. Common Market Studies\/}~{\em 50\/}(4), 670--683.

\bibitem[\protect\citeauthoryear{Slapin and Proksch}{Slapin and
  Proksch}{2010}]{Slapin:2010il}
Slapin, J.~B. and S.~O. Proksch (2010).
\newblock {Look who's talking: Parliamentary debate in the European Union}.
\newblock {\em European Union Politics\/}~{\em 11\/}(3), 333--357.

\bibitem[\protect\citeauthoryear{Stevens, Kegelmeyer, Andrzejewski, and
  Buttler}{Stevens et~al.}{2012}]{stevens12coherence}
Stevens, K., P.~Kegelmeyer, D.~Andrzejewski, and D.~Buttler (2012).
\newblock Exploring topic coherence over many models and many topics.
\newblock In {\em Proceedings of the 2012 Joint Conference on Empirical Methods
  in Natural Language Processing and Computational Natural Language Learning},
  pp.\  952--961. Association for Computational Linguistics.

\bibitem[\protect\citeauthoryear{Steyvers and Griffiths}{Steyvers and
  Griffiths}{2006}]{steyvers06prob}
Steyvers, M. and T.~Griffiths (2006).
\newblock {Probabilistic Topic Models}.
\newblock In T.~Landauer, D.~Mcnamara, S.~Dennis, and W.~Kintsch (Eds.), {\em
  Latent Semantic Analysis: A Road to Meaning.} Laurence Erlbaum.

\bibitem[\protect\citeauthoryear{Sulo, Berger-Wolf, and Grossman}{Sulo
  et~al.}{2010}]{sulo10temporal}
Sulo, R., T.~Berger-Wolf, and R.~Grossman (2010).
\newblock Meaningful selection of temporal resolution for dynamic networks.
\newblock In {\em Proc. 8th Workshop on Mining and Learning with Graphs}, pp.\
  127--136. ACM.

\bibitem[\protect\citeauthoryear{Wang, Cao, Xu, and Li}{Wang
  et~al.}{2012}]{wang12group}
Wang, Q., Z.~Cao, J.~Xu, and H.~Li (2012).
\newblock Group matrix factorization for scalable topic modeling.
\newblock In {\em Proc. 35th SIGIR Conf. on Research and Development in
  Information Retrieval}, pp.\  375--384. ACM.

\end{thebibliography}
